\pdfoutput=1

\documentclass[11pt]{article}

\usepackage[final]{acl}

\usepackage{times}
\usepackage{latexsym}

\usepackage[T1]{fontenc}

\usepackage[utf8]{inputenc}

\usepackage{microtype}

\usepackage{inconsolata}

\usepackage{graphicx}
\usepackage{amsmath}
\usepackage{multirow}

\usepackage{algorithm}
\usepackage{algorithmic}
\usepackage{booktabs} 
\usepackage{xcolor}
\newcommand{\modelname}{ChemActor}
\usepackage{colortbl} 
\definecolor{mygray}{gray}{.9}

\usepackage[symbol]{footmisc} 
\makeatletter
\def\@fnsymbol#1{%
  \ifcase#1
  \or {$\dag$}
  \or {$\ast$}
  \else \@ctrerr
  \fi
}
\makeatother

\title{ChemActor: Enhancing Automated Extraction of Chemical \\ Synthesis Actions with LLM-Generated Data}

\author{
 \textbf{Yu Zhang\textsuperscript{1}\thanks{Equal contribution. }}, 
 \textbf{Ruijie Yu\textsuperscript{1}\footnotemark[1]},
 \textbf{Jidong Tian\textsuperscript{1}},
 \textbf{Feng Zhu\textsuperscript{2}},
\\
 \textbf{Jiapeng Liu\textsuperscript{3}},
 \textbf{Xiaokang Yang\textsuperscript{1}},
 \textbf{Yaohui Jin\textsuperscript{1}\thanks{Corresponding authors. }},
 \textbf{Yanyan Xu\textsuperscript{1}\footnotemark[2]},
\\
 \textsuperscript{1} AI Institute, Shanghai Jiao Tong University, China \\
 \textsuperscript{2} Frontiers Science Center for Transformative Molecules, Shanghai Jiao Tong University, China \\
 \textsuperscript{3} X-Imaging Intelligent Technology (Shanghai) Co. LTD., China \\
 \small{\texttt{\normalsize{\{cynthiazhang, el\_iauk, frank92, fzchem, xkyang, jinyh, yanyanxu\}@sjtu.edu.cn}}}\\
 \small{{\texttt{\normalsize{liujiapeng@x-imaging.com}}}}
}

\begin{document}
\maketitle


\begin{abstract}
With the increasing interest in robotic synthesis in the context of organic chemistry, the automated extraction of chemical procedures from literature is critical. However, this task remains challenging due to the inherent ambiguity of chemical language and the high cost of human annotation required for developing reliable computer-aided extraction protocols. Here, we present \textbf{ChemActor}, a fully fine-tuned large language model (LLM), as a chemical executor to convert between unstructured experimental procedures and structured action sequences. We propose a sequential LLM-generated data framework to address the challenges of insufficient and low-quality annotated data. This framework integrates a data selection module that selects data based on distribution divergence, with a general-purpose LLM, to generate machine-executable actions from a single molecule input. Additionally, we introduce a novel multi-round LLMs circle review metric, which reflects the model's advanced understanding of chemical experimental procedures. Extensive experiments on reaction-to-description (R2D) and description-to-action (D2A) tasks demonstrate that ChemActor, augmented by LLM-generated data, achieves state-of-the-art performance, outperforming the baseline model by 10\%. The code is available at: \href{https://github.com/Zhanghahah/ChemActor}{https://github.com/Zhanghahah/ChemActor}.  

\end{abstract}

\section{Introduction}


In the field of organic synthesis, robotic systems for chemical synthesis have grown in popularity. The availability of structured chemical data becomes more important owing to its potential to accelerate the discovery of transformative molecules with minimal human intervention~\citep{godfrey2013remote, peplow2014robo, guo2021automated, canty2024reproducibility}. The execution of chemical synthesis necessitates a thorough extraction of the precise sequence of procedures, including addition, stirring, and concentration, as well as the optimal parameters for these actions, such as temperature and atmosphere~\citep{walker2019learning, fan2024openchemie}, etc. However, chemical experimental procedures consist of various writing styles in natural language~\citep{zhang2024fine}, and extraction of experimental procedures from literature requires manual revisions~\citep{mehr2020universal}. Although chemists have focused on designing reliable tools through natural language processing~\citep{guo2021automated}, quite often these solutions rely on identifying and utilizing rules specific to each data item at the cost of a substantial human effort~\citep{davies2019digitization, vaucher2020automated, vaucher2021inferring, zeng2023transcription}.

Nowadays, the emergence of generative pre-trained transformer-based large language models (LLMs), typified by GPT-4, has sparked significant interest in the field of AI for chemistry~\citep{baum2021artificial, boiko2023autonomous, m2024augmenting}. Pre-trained on a vast corpus of chemical reactions literature, LLMs are endowed with fundamental chemical knowledge through text-to-text generation~\citep{achiam2023gpt}. However, the sparsity of chemical data and the lack of high-quality annotations still limit the development of applying LLMs to autonomous chemical experiments~\citep{coley2019robotic}. Due to the ambiguity and diverse writing styles in the chemical literature~\citep{zhang2024fine}, extraction of experimental procedures from literature by LLMs often requires manual revisions~\citep{zhong2023reactie}. This presents a significant challenge for converting chemical descriptions into machine-readable annotated experimental actions~\citep{ji2023survey,zhang2023siren}. Overall, \textit{the design of an automated conversion from unstructured chemical descriptions into structured ones for LLM-based autonomous chemical experiments is a desirable and needed technology.}

Recently, the technology of LLM-generated data has been very popular as it can enhance the model's own performance by generating valuable training data~\citep{ li2024synthetic}. Several studies, particularly in vision understanding and nature language fields~\citep{tremblay2018training}, have highlighted that generated data augmentation markedly improves the performance of foundational models~\citep{gao2023synthetic, trinh2024solving, li2024synthetic}. Considering that the standard of the recording and subsequent reporting of the synthesis of new reactions varies greatly, we investigate a naturally arisen question: \textit{can scaling up high-quality LLM-generated data enhance the performance of the LLMs for generating experimental action sequences tasks?}
To answer this, we propose a novel LLM-generated data framework into ChemActor for generating machine-executable actions starting from a molecule. The reason why LLM-generated data can be a potential solution is that (i) via joint learning with generated data and real data, LLMs learn relationships between chemical synthetic descriptions and action sequences, thereby acquiring chemical knowledge akin to the learning process of chemists; (ii) the data selection module generates more valuable data that compensates for the uneven distribution of chemical space, thereby enhancing the representation of reaction description.
The contributions of this work can be summarized as follows:
 \begin{enumerate}
     \item  We propose a 7B-scale fine-tuned LLM, a.k.a. \modelname{}, as a chemical executor to convert unstructured human-readable descriptions to structured machine-executable actions.  
     \item We design a sequential LLM-generated data framework that integrates a data selection module--based on distribution divergence--with a general-purpose LLM to generate actions from a single molecule input.
     \item  We design a multi-round LLMs circle review metric, using interactive debate prompting to prove \modelname{}'s advanced semantic understanding of chemical experimental knowledge.
      
 \end{enumerate}

\section{Related Work}

\paragraph{Automated Extraction of Chemical Synthesis Actions} 
Chemical experimental actions extraction aims at extracting structural information, which is prepared for an automated synthesis platform. According to different sources, this task includes two categories of sub-tasks: description-to-action extraction (D2A) and reaction-to-action extraction (R2A). D2A is a natural language information extraction task that converts unstructured experimental descriptions to structured synthetic steps. Vaucher et al.~\citep{vaucher2020automated} first propose the D2A task with pre-defined synthesis actions and design a pre-trained Transformer-based model to solve it. Zeng et al.~\citep{zeng2023transcription} introduce a comprehensive schema for D2A, and also design a new dataset, \textsc{ChemTrans}. Meanwhile, they propose knowledge-enhanced methods for fine-tuned models (T5) and large language models (GPT-3.5), respectively. Results show that fine-tuned models with chemical knowledge exhibit better performance on D2A. R2A is to predict chemical experimental actions directly from chemical equations. Vaucher et al.~\citep{vaucher2021inferring} propose the first R2A task that converts reaction SMILES into actions. Meanwhile, they also design a series of Transformer-based models to solve R2A. Liu et al.~\citep{liu2024reactxt} construct a new R2A dataset, \textsc{OpenExp}, and propose a novel incrementally pre-trained language model, ReactXT, to achieve such a task. Compared with other general or scientific language models, ReactXT shows a significant improvement on \textsc{OpenExp}. All these studies have provided effective datasets and baseline models, but D2A and R2A remain to be addressed. In this work, we introduce a novel LLM-generated augmentation framework that can effectively enhance the pre-trained models' ability for the extraction of chemical experimental actions.

\paragraph{LLM-Generated Data}
In recent years, data generation technologies have transformed data science by mimicking real-world data, addressing challenges of data scarcity~\citep{wang2024llm}. Several studies, particularly in mathematical and medical fields, have highlighted that realistic data augmentation markedly improves the performance of machine learning models~\citep{trinh2024solving,gao2023synthetic}. Recently, LLMs have exhibited the ability to generate highly reliable data~\citep{li2023synthetic,li2024synthetic}. These methods are widely used in the field of natural language processing~\citep{chen2023empirical}. To be specific, Whitehouse et al.~\citep{whitehouse2023llm} utilize pre-trained LLM, such as GPT-4, to generate more data for low-resource machine translation. Cai et al.~\citep{cai2023resolving} propose a generative framework that uses LLaMA to resolve the data imbalance problem. Yuan et al.~\citep{yuan2023large} take advantage of LLM to enhance the compatibility of clinical trial descriptions. These studies provide a foundation for LLM-generated data methods, but applying them to the scientific domain is still challenging. This is because pre-trained LLMs do not necessarily possess the corresponding domain knowledge.

\section{Methods}
\subsection{Problem Setup}
Given an unstructured reaction description text $\mathbf{d}$ from the chemical literature, our objective is to extract all action sequences $\mathbf{a}$.

\begin{figure*}[t]
\centering 
\includegraphics[width=1\textwidth]{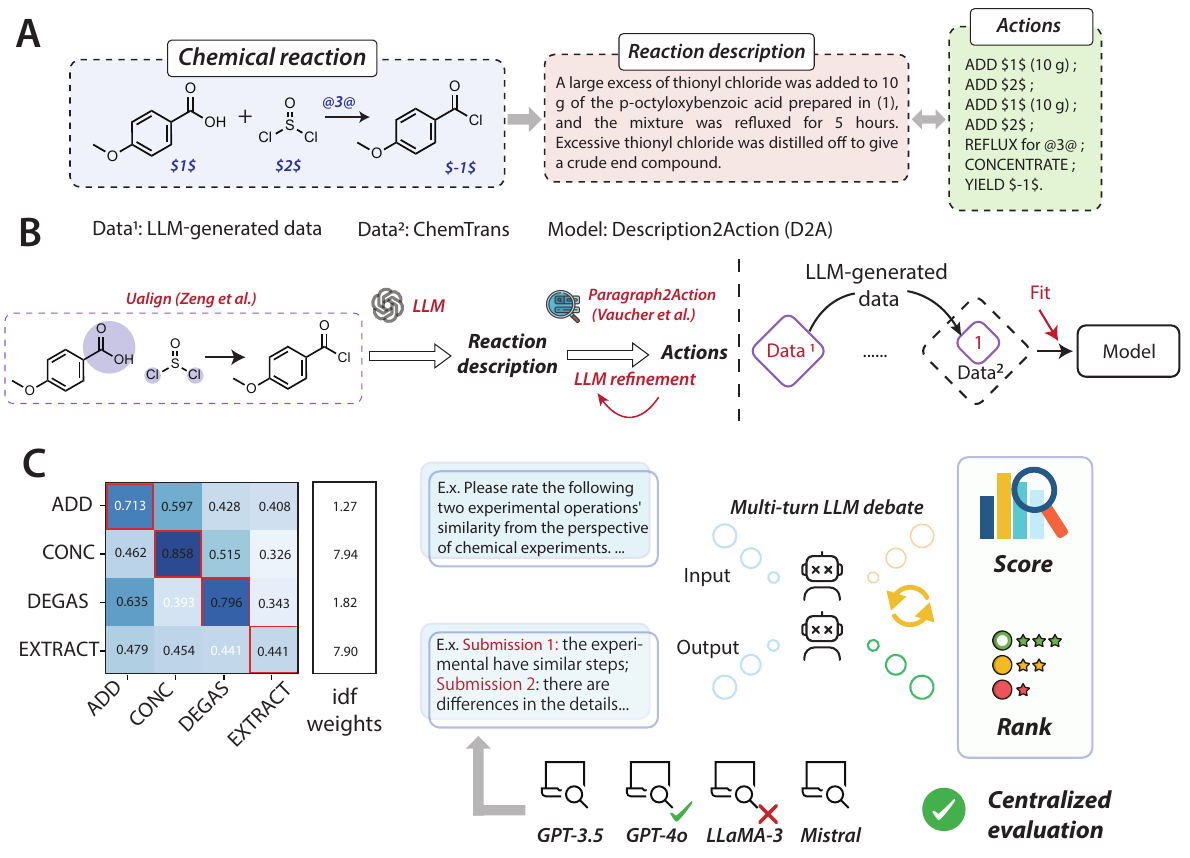}
\caption{The overview of \modelname{}. \textbf{(A)} Task definition; \textbf{(B)} Illustration of the framework of LLM-generated data; \textbf{(C)} BERTScore and multi-round LLMs circle review metrics for evalution.}
\label{fig:framework}
\end{figure*}

\subsection{The Framework of LLM-generated Data}
Here, we first introduce a novel LLM-generated data framework for generating machine-executable actions starting from a single molecule. The overview is shown in Figure~\ref{fig:framework}. In Figure~\ref{fig:framework}A, we start with a sampled chemical reaction $\mathbf{r}$, along with a detailed description and its experimental actions. Firstly, we design a comprehensive data template to construct pair-wised instructed Q\&A datasets for supervised fine-tuning based on the original LLaMA-2 foundational model. The detailed data construction is illustrated in the Appendix~\ref{sec: data illu} and Figure~\ref{fig:data-example}. Subsequently, we obtain a reaction-to-description model, a.k.a. \textbf{R2D}, and a description-to-action model, a.k.a. \textbf{D2A}, repspectively. \textbf{R2D} converts $\mathbf{r}$ into reaction descriptions $\mathbf{d} = \{d_1,d_2,\cdots,d_n\}$. The objective of \textbf{R2D} is shown in Equation~\ref{eq1}, where $\theta$ is the trainable parameters, $d_{<i}$ represents $\{d_0, \cdots, d_{i-1}\}$ tokens, and $d_0$ is the starting token. \textbf{D2A} model focuses on converting reaction descriptions into a series of corresponding experimental actions $\mathbf{a} = \{a_1,\cdots,a_m\}$, such as adding, refluxing, etc. The objective of \textbf{D2A} are provided in Equation~\ref{eq2}, where $\psi$ is the trainable parameters, $a_{<i}$ represents $\{a_0, \cdots, a_{i-1}\}$, and $a_0$ is the starting token.

In Figure~\ref{fig:framework}B, we present the LLM-generated data framework, which begins with a target molecule. First, we employ an efficient single-step retrosynthesis model from previous work~\citep{zeng2024ualign} to generate the necessary reactants. To ensure the validity of the generated reaction, we further utilize a forward prediction model~\citep{schwaller2020predicting} for self-correctness. Next, we leverage a general-purpose LLM, GPT-4o, to generate reaction descriptions using an in-context learning strategy, enhancing generation accuracy through context augmentation with relevant examples. Subsequently, the Paragraph2Actions model~\citep{vaucher2020automated} is applied to predict procedural actions, and general-purpose LLM, GPT-4o, is utilized for formatting refinement. During training, we augment the existing datasets with LLM-generated data to enhance \modelname{}’s capability in translating reaction descriptions into procedural actions.
\begin{algorithm}[h]

\caption{\textbf{Data Selection Module}}
\label{alg1}
\begin{algorithmic}
\STATE \textbf{Input:} triplet set $O=[(r^*_1,d^*_1,a^*_1), (r^*_2,d^*_2,a^*_2),...(r^*_k,d^*_k,a^*_k)]$ \\ of LLM-generated data, the encoder of R2D model $f_\theta^{enc}$, the encoder of D2A model $f_\psi^{enc}$, distribution difference $\delta$, the original input data points of reaction descriptions $\mathbf{d}$, the LLM-generated data points $\mathbf{d^*}$, threshold of distribution difference $\tau$, total number of LLM-generated data $n$.
\STATE \textbf{Output:} selected LLM-generated data sets $A$.
\STATE \textbf{Initialization:} $A = \emptyset$
\STATE \textbf{Initialization:} $i = 0$
\WHILE{$i * k < n$}
    \STATE $\mathbf{e} = [e_1, e_2, \cdots, e_k] = f_\psi^{enc}(\mathbf{d})$
    \STATE $\mathbf{e^*} = [e^*_1, e^*_2, \cdots,e^*_k] = f_\psi^{enc}(\mathbf{d}^*)$
    \STATE $\delta =  \lvert KL(\mathbf{e})-KL(\mathbf{e^*}) \rvert$
    \IF{$\delta \geq \tau$}
        \STATE $A.{\rm append}(O)$
        \STATE$i = i+1$
    \ENDIF
\ENDWHILE
\RETURN $A$
\end{algorithmic}
\end{algorithm}

\subsection{Data Selection Module}
We further design the data selection module to enhance the model performance, as outlined in Algorithm~\ref{alg1}. First, we extract the embeddings of original datasets by \textbf{D2A} model mentioned in Figure~\ref{fig:framework}A, denoted as $\mathbf{e} = [e_1, e_2, \cdots, e_k]$. We randomly select the triplet set $k$-consisting of data (reaction formulation, description, actions) from LLM-generated data sets, and extract its embeddings by \textbf{D2A}, denoted as $\mathbf{e^*} = [e^*_1, e^*_2, \cdots,e^*_k]$. Next, we project the embeddings of both real and LLM-generated data into a two-dimensional representation space using the uniform manifold approximation and projection (UMAP) algorithm. To ensure the quality of the LLM-generated data, we employ a selection criterion based on the Kullback-Leibler (KL) divergence between the distributions of LLM-generated and real data in this low-dimensional space. Specifically, we compute the KL divergence for the distributions of LLM-generated data and original data, denoted as $KL(e^*)$, and $KL(e)$, respectively. We then define a threshold $\tau$ for the difference $\delta =  \lvert KL(\mathbf{e})-KL(\mathbf{e^*}) \rvert$. The selection process operates as follows: if $\tau > \delta$, the corresponding triplet sets from the LLM-generated dataset are retained; otherwise, the data are resampled. This iterative procedure continues until a predefined number $n$ of qualified LLM-generated samples is obtained.

\begin{equation}\label{eq1}
    \max_{\theta} \; \log \; P_\theta(\mathbf{d}|\mathbf{r}) = \sum_{i=1}^{n} \;  \log \; P_\theta(d_i|\mathbf{r}, d_{<i})
\end{equation}
\begin{equation}\label{eq2}
    \max_{\psi} \; \log \; P_\psi(\mathbf{a}|\mathbf{d}) = \sum_{i=1}^{m} \;  \log \; P_\psi(a_i|\mathbf{d}, a_{<i})
\end{equation}

\subsection{Semantic Evaluation for Actions}
To better evaluate \modelname{}, we introduce two additional metrics: BERTScore~\citep{zhang2019bertscore} and LLMs circle review, as shown in Figure~\ref{fig:framework}C. The detailed introduction of the metric of BERTScore can be found in the Appendix~\ref{app:metric}. Further, LLM-based peer review presents a promising approach for evaluating generated outputs~\citep{chu2024pre, zhao2024auto}. Inspired by~\citep{chern2024can}, we design an LLMs circle review metric based on a multi-round debating strategy to better evaluate \modelname{}. Specifically, we implement an interactive evaluation framework employing four LLMs (GPT-3.5-turbo, GPT-4, LLaMA-3, and Mistral) to independently assess prediction-annotation semantic similarity. Through iterative prompting, these LLMs transparently share scores and rationales, then engage in structured debate. The debate continues until a predetermined number of rounds is reached, after which the final average score is adopted. The prompts for multi-round debating are shown in Figure~\ref{fig:debate}.

\section{Experiment}
\subsection{Data}

We evaluate the effectiveness of our method using two datasets, \textsc{ChemTrans} and \textsc{OpenExp}. \textsc{ChemTrans}~\citep{zeng2023transcription} consists of 3,950 description–action pairs with a detailed schema of instructions. \textsc{OpenExp} proposed by~\citep{liu2024reactxt} are derived from USPTO-Applications~\citep{lowe1976chemical} and the Open Reaction Database~\citep{kearnes2021open}. 
Further, Appendix~\ref{A: dataset} provides the data statistics and distribution of these datasets. 


\begin{table*}[htbp]
\centering
\small
\renewcommand\arraystretch{1.08}
\resizebox{\linewidth}{!}{

\setlength{\tabcolsep}{1.4mm}{\begin{tabular}{l|ccccc|cccc}
\toprule
\textbf{Task}& \multicolumn{5}{|c}{\textbf{Description-to-Action (D2A)}} & \multicolumn{4}{|c}{\textbf{Action-to-Description (A2D)}} \\
\midrule
\textbf{Model} & \textbf{SM-A} & \textbf{SM-O} & \textbf{BLEU-2} & \textbf{BLEU-4} & \textbf{EM} & \textbf{Distinct-4} & \textbf{ROUGE-4} & \textbf{BLEU-2} & \textbf{BLEU-4} \\
\midrule
Paragraph2Actions & 21.57 & 57.91 & 44.88 & 27.19 & 0 & 8.308 & 0.517 & 5.210 & 0.365 \\
Paragraph2Actions+ & 22.43 & 58.45 & 44.97 & 27.97 & 0 & 18.36 & 1.225 & 8.168 & 0.933 \\
GPT-3.5 & 0.441 & 4.471 & 7.520 & 0.931 & 0 & 67.99 & 5.261 & 10.83 & 2.920 \\
GPT-3.5, 3-shot & 37.53 & 66.96 & 59.69 & 44.91 & 4.94 & 56.51 & 13.39 & 20.41 & 8.816 \\
GPT-3.5, 3-shot* & 45.11 & 70.45 & 62.84 & 50.16 & 6.71 & 59.26 & 15.06 & 23.19 & 10.69 \\
\midrule
GPT-3.5-chat & 2.708 & 35.49 & 14.17 & 2.72 & 0 & \textbf{74.62} & 3.016 & 6.423 & 1.982 \\
GPT-3.5-chat, 3-shot & 25.75 & 57.99 & 49.25 & 31.92 & 0.72 & \underline{70.70} & 8.619 & 15.73 & 5.486 \\
GPT-3.5-chat, 3-shot* & 34.88 & 62.28 & 55.57 & 40.45 & 3.25 & 69.59 & 10.33 & 17.96 & 6.913 \\
\midrule
T5-ChemTrans, base & 65.59 & 83.78 & 59.24 & 43.46 & 18.31 & 56.67 & 20.06 & 27.61 & 15.01 \\
T5-ChemTrans, large & 67.12 & \underline{85.41} & \underline{75.89} & 67.33 & 22.36 & 57.17 & \underline{21.82} & 29.54 & 16.55 \\
\midrule
\textbf{\modelname{}}, wo/generated & \underline{68.29} & 85.07 & 75.63 & \underline{68.37} & \underline{31.90} & 56.71 & 19.18 & \underline{29.87} & \underline{18.43} \\
\textbf{\modelname{}}, w/generated & \textbf{76.88} & \textbf{89.01} & \textbf{84.74} & \textbf{76.93} &  \textbf{36.40} & 64.19 & \textbf{25.79} & \textbf{31.76} & \textbf{34.27} \\
\bottomrule
\end{tabular}}
}
\caption{Experimental Results of ChemActor on \textsc{ChemTrans} dataset. The best performance is in \textbf{bold.}}
\label{tab:performance-chemtrans}
\end{table*}


\subsection{Experimental Setup}
Firstly, we organize training data as instruction prompt datasets for supervised fine-tuning. For the training strategy, we propose an alternating data mixing paradigm. This approach employs an iterative training scheme where the model is first exposed to a batch of real data, followed by a batch of LLM-generated data. Fully fine-tuning LLaMA-2-7B for 8 epochs is completed in approximately 12 hours using 8 $\times$ NVIDIA A800 GPUs. We set $\tau$ as 0.7 in data selection module, the batch size to 12, and the default learning rate to 9.65e-6. The further detailed training setting can be seen in Appendix~\ref{app:hyper}.

\subsection{Performance Comparison}
We evaluate baseline and \modelname{} methods on the two benchmark datasets, \textsc{ChemTrans} and \textsc{OpenExp}, respectively. Descriptions of metrics applied for evaluation and introduction of all baseline methods are introduced in Appendix~\ref{app:metric}.


\begin{table*}[htb]
\centering
\small
\renewcommand\arraystretch{1.08}
\resizebox{\linewidth}{!}{

\setlength{\tabcolsep}{1.5mm}{\begin{tabular}{l|ccc|cccc|ccc}
\toprule

\textbf{Method} & \textbf{\scriptsize Validity } & \textbf{\scriptsize BLEU-2} & \textbf{\scriptsize BLEU-4} & \textbf{\scriptsize 100\%LEV} & \textbf{\scriptsize 90\%LEV} & \textbf{\scriptsize 75\%LEV} & \textbf{\scriptsize 50\%LEV} & \textbf{\scriptsize ROUGE-1} & \textbf{\scriptsize ROUGE-2} & \textbf{\scriptsize ROUGE-L} \\
\midrule
\rowcolor{cyan!10}
\multicolumn{11}{l}{\textbf{\textit{With molecule pre-training, reaction equation $\to$ experimental actions}}} \\
TextChemT5$_{220M}$                & 99.3 & 54.1 & 40.6 & 0.4 & 4.6 & 13.7 & 61.2 & 61.5 & 40.3 & 56.4 \\
MolT5-Large$_{780M}$               & 99.6 & 54.5 & 41.0 & 0.6 & 6.6 & 16.6 & 63.7 & 62.5 & 40.9 & 57.2 \\
Galactica$_{1.3B}$                 & 99.9 & 53.5 & 39.5 & 0.4 & 5.7 & 13.4 & 60.5 & 60.9 & 38.6 & 55.2 \\
MolCA, Galac$_{1.3B}$              & 99.9 & 54.9 & 41.5 & 1.0 & 9.2 & 18.9 & 65.3 & 62.5 & 40.4 & 57.0 \\
ReactXT, Galac$_{1.3B}$      & \textbf{100.0} & \underline{57.4} & \underline{44.0} & \underline{1.0} & \textbf{9.5} & \underline{22.6} & \underline{70.2} & \underline{64.4} & \underline{42.7} & \underline{58.9} \\
\midrule
\textbf{ChemActor*}, wo/g     & \textbf{100.0} & \textbf{86.7} & \textbf{85.4} & \textbf{2.0} & \underline{7.0} & \textbf{78.5} & \textbf{99.0} & \textbf{86.7} & \textbf{85.3} & \textbf{86.1} \\
\midrule

\rowcolor{cyan!10}
\multicolumn{11}{l}{\textbf{\textit{Without molecule pre-training, reaction description $\to$ experimental actions}}} \\
Paragraph2Actions & 63.2 & 34.5 & 19.1 & 0.0 & 0.0 & 0.0 & 13.6 & 46.6 & 18.1 & 36.4 \\
Paragraph2Actions+                   & 76.0 & 45.0 & 30.7 & 0.6 & 6.5 & 13.0 & 38.4 & 55.7 & 29.2 & 47.0 \\
T5-ChemTrans, base & 99.5 & \underline{92.2} & \underline{90.4} & \underline{38.7} & \textbf{69.6} & \textbf{91.6} & \underline{99.0} & \textbf{95.1} & \textbf{93.6} & \textbf{94.6} \\
\midrule
 \textbf{\modelname{}}, wo/g & \textbf{99.8} & \textbf{92.9} & \textbf{91.0} & \textbf{40.1} & \underline{69.4} & \underline{91.4} & \textbf{99.2} & \underline{92.5} & \textbf{93.6} & \underline{94.5} \\
\bottomrule
\end{tabular}}
}
\caption{Experimental Results of ChemActor on \textsc{OpenExp} dataset. The best performance is in \textbf{bold.}}
\label{tab:performance-openexp}
\end{table*}

\textbf{Evaluation on \textsc{ChemTrans}.} We compare \modelname{} with various existing methods, as presented in Table~\ref{tab:performance-chemtrans}. Additionally, we evaluate the performance of general-purpose model GPT-3.5 and GPT-3.5-chat API for comparison, which can be seen in the 
Appendix Figure~\ref{fig:baseline-result}. Following previous work~\citep{zeng2023transcription}, we assess performance using the Sequence Matching (SM) and ExactMatch (EM) metrics.

In Table \ref{tab:performance-chemtrans}, \textbf{ChemActor, w/generated} refers to ChemActor trained with an additional 50,000 LLM-generated data points, while \textbf{ChemActor, w/o generated} represents \modelname{} trained without LLM-generated data. We evaluate the performance of our \modelname{} for both D2A and A2D tasks. From the results, we conclude that (i) the GPT-3.5 series, even with few-shot instructions, struggles with translating between human-readable synthetic descriptions and machine-executable actions, leading to low similarity metrics; (ii) \modelname{} wo/generated outperforms both T5-ChemTrans and Paragraph2Actions\citep{vaucher2021inferring}, achieving 68.37\% BLEU-4 and 31.9\% EM scores for the D2A task, representing improvements of 41.18\% and 31\%, respectively. Furthermore, by incorporating LLM-generated data into the training process, \modelname{}, w/generated achieves a 9\% improvement in BLEU-2 and a 4.5\% increase in EM compared to \modelname{}, w/o generated for generating action sequences. The improvement can be attributed to the incorporation of LLM-generated data, which enriches the model’s understanding of the relationship between reaction descriptions and corresponding experimental actions, thereby enhancing its ability to generate accurate action sequences. Furthermore, we conduct a type-matching experiment to evaluate the model's ability to predict action types and their corresponding necessary components, as shown in Appendix~\ref{app:type matching}.

\textbf{Evaluation on \textsc{OpenExp}.} For \textsc{OpenExp}, we compare our \modelname{} with the state-of-the-art scientific LLMs and comparative baseline methods. Notably, all these scientific LLMs are pre-trained on molecules using 1D or 2D SMILES representations rather than reaction descriptions. They take the reaction equation as input and generate experimental action sequences. In contrast, the baselines proposed by~\citep{vaucher2021inferring} and ~\citep{zeng2023transcription} use reaction descriptions as input and generate structural experimental actions.


Following~\citep{vaucher2021inferring, liu2024reactxt}, we employ the additional metrics for performance evaluation: Validity and  Levenshtein score (LEV).
The performance of our \modelname{} on \textsc{OpenExp} datasets is shown in Table~\ref{tab:performance-openexp}. The results reveal that \modelname{}, wo/generated consistently outperforms baseline methods across all metrics. Specifically, it surpasses ReactXT by 34.2\% for BLEU-2 and 68.5\% for 75\%LEV, demonstrating \modelname{}'s effectiveness for text-based reaction understanding. Furthermore, baseline models that use reaction equations as input consistently underperform compared to those that leverage reaction descriptions. This is because reaction descriptions contain detailed contextual and procedural information about experimental conditions, which are essential for accurate predictions. In contrast, reaction equations primarily represent reactants and products, lacking explicit details about reaction conditions, solvents, catalysts, and procedural steps that significantly influence outcomes.

\subsection{Multi-Round Circle Review}
As previously discussed, existing metrics are insufficient for evaluating the semantic correctness of predicted outputs. Thus, we propose BERTScore and LLMs circle review for further evaluation. The details of the proposed metric are illustrated in Appendix Section~\ref{app:metric}. We also perform a human evaluation experiment to substantiate the effectiveness of our proposed metric. Specifically, we invite six Ph.D. and M.S. students from the department of chemistry to assess the prediction results on the \textsc{ChemTrans} test set. Each prediction is required to be scored on a scale from 0 to 10. The entire test set is divided into six groups, and each student is responsible for evaluating the cases in their assigned group. Next, we collect all scores, average them, and scale them to a range of 0 to 1 to obtain the final evaluations. 

\begin{table*}[!ht]
\centering
\small
\renewcommand\arraystretch{1.08}

\setlength{\tabcolsep}{1mm}{\begin{tabular}{l|cccc|c}
\toprule
\textbf{Model} & \textbf{GPT-3.5} & \textbf{GPT-4o} & \textbf{LLaMA-3} & \textbf{Mistral} & \textbf{Human review} \\
\midrule
Mistral & 0.74 & 0.76 & 0.78 & 0.77 & 0.61 \\
T5-Base & 0.68 & 0.70 & 0.75 & 0.70 & 0.50 \\
T5-ChemTrans & 0.76 & 0.78 & \textbf{0.81} & \textbf{0.78} & 0.58 \\
ChemActor (wo/g) & \textbf{0.77} & \textbf{0.79} & \textbf{0.81} & \textbf{0.78} & \textbf{0.66} \\
\bottomrule
\end{tabular}}
\caption{Results for multi-round LLMs circle reviews and human evaluations on \textsc{ChemTrans} dataset. The best performance is in \textbf{bold.}}
\label{tab:circle-evaluation-human}
\end{table*}

In Table~\ref{tab:circle-evaluation-human}, evaluation results reveal the insights: 1) within the LLM circle review metric, all general-purpose LLMs consistently achieve the same ranking across all compared methods: ChemActor>T5-ChemTrans>Mistral>T5-Base. The superior performance of T5-ChemTrans over Mistral can be attributed to its pre-training on literature data encompassing experimental actions, which significantly enhances its understanding of chemical knowledge. In contrast, T5-Base and Mistral-7B are fine-tuned using LoRA from general foundation models, making it challenging for them to perform well with the limited \textsc{ChemTrans} training datasets. 2) Human evaluation scores tend to be conservative, as they are all lower than the scores from the LLM circle review. Importantly, the ranking trend of the human evaluations is consistent with the results from the LLM circle review. We hypothesize that the reason Mistral performs better than T5-ChemTrans in this context is that Mistral's generated results are more closely aligned with human preference since it is pre-trained on a massive, unlabeled text corpus.

In addition, we conduct multi-round LLM debate experiments and calculate the scores of methods to verify the effectiveness of our proposed method. Detailed prompt examples can be seen in Appendix Figure~\ref{fig:debate}. In Figure~\ref{fig:peer review}, it shows that ChemActor outperforms other baselines on both BERTScore and LLMs circle review scores. Both ChemActor and T5-ChemTrans achieve BERTScore F1 scores exceeding 0.85, with ChemActor obtaining the highest score, highlighting its superior ability to capture the semantics of chemical procedures. Although circle review scores generally align with the BERTScore, there are variations between models. Specifically, GPT-4o and LLaMA-3 rate ChemActor the highest, whereas GPT-3.5 and Mistral-7B favor T5-ChemTrans. 
\begin{figure}[t]
\centering 
\includegraphics[width=0.5\textwidth]{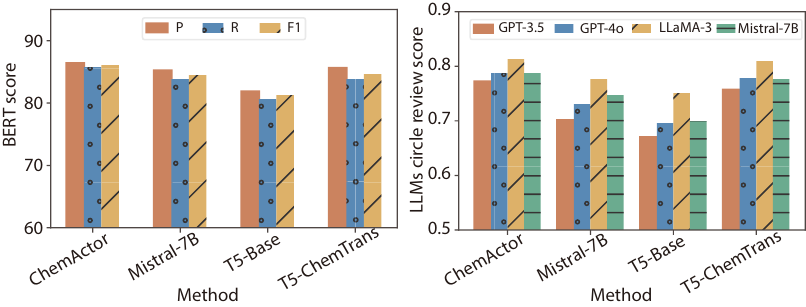}
\caption{Results for multi-round LLMs circle reviews.}
\label{fig:peer review}
\end{figure}
\subsection{Ablation Study}

LLMs show increasingly advanced emergent capabilities and are being incorporated across various domains. Understanding the relationship between the LLM-generated data and its abilities holds significant importance. Thus, we design several ablation studies to verify the effectiveness of our proposed framework.
\subsubsection{Data Selection Module}
The data selection module employs a threshold $\tau$ to filter LLM-generated data based on distribution divergence. Here we discuss the effect of hypereparameter $\tau$ for model performance.
\begin{table*}[ht!]
\centering
\small
\renewcommand\arraystretch{1.08}

\setlength{\tabcolsep}{2mm}{\begin{tabular}{c|c|c|cccccc}
\toprule

\textbf{Model}    & \textbf{$\tau$}    & \textbf{Projection}          & \textbf{SM-A} & \textbf{SM-O} & \textbf{BLEU-2} & \textbf{BLEU-4} & \textbf{EM} & \textbf{Cost Time}  \\ 
\midrule

\multirow{4}{*}{ChemActor, w/g}& 0.5 & UMAP &67.13  & 83.31 & 74.03  &66.87 &30.94 & 141s \\
& 0.6  &UMAP & 68.29 & 84.97 & 76.17   & 68.74  & 32.41 &  385s \\ 
& 0.7   &UMAP & 68.92 & 85.51 & 76.01 & 68.71& 34.14   &  427s \\ 
& 0.8   &UMAP & 68.93 & 85.53 & 76.07 & 69.12 & 34.33 &  1031s \\ 
\midrule
\multirow{1}{*}{ChemActor, w/g} & 0.7  &t-SNE  & 66.86 & 83.09 & 73.79 & 66.81 & 29.89 &  517s \\ 

\bottomrule

\end{tabular}} 
\caption{Performance evaluation of ChemActor with varying amounts of $\tau$ on the \textsc{ChemTrans} dataset. Cost time refers to the time spent searching for LLM-generated data points that fulfill the requirements.}
\label{tab: data selection module}
\end{table*}
The results in Table~\ref{tab: data selection module} demonstrate that ChemActor's performance is significantly influenced by both the divergence threshold $\tau$ and the dimension reduction method. Increasing $\tau$ from 0.5 to 0.8 consistently improves all evaluation metrics. However, this improvement comes at a computational cost, as processing time increases dramatically from 141s to 1031s.
The comparison between the UMAP and the t-SNE reveals that UMAP outperforms t-SNE across all metrics. It is indicated that as $\tau$ increases, the spatial distribution of sampled points becomes more uniform, which helps mitigate the issue of data skewness for training. Further, the results demonstrate that $\tau=0.7$ with UMAP projection represents an optimal balance between model performance and computational efficiency.


\subsubsection{Training Size Proportions}
Before investigating the effect of LLM-generated data on performance, we examine whether increasing the training data can improve the performance of \modelname{}. To this end, we evaluate \modelname{} with varying proportions of training data on the \textsc{ChemTrans} and \textsc{OpenExp} datasets. We also introduce T5-Base~\citep{raffel2020exploring}, LLaMA-3, and Mistral-7B~\citep{jiang2023mistral} for comparison. Note that the T5-Base refers that we use the T5 model proposed by \citep{raffel2020exploring} for fine-tuning. The entire five scenarios for evaluation are illustrated in Appendix~\ref{app: training size}. We list four training set proportions, 30\%, 50\%, 70\%, and 100\% in Table~\ref{tab:ChemTrans} and Table~\ref{tab:openexp}. 
For the \textsc{ChemTrans} dataset, it is notable that Mistral-7B consistently outperforms T5-Base and LLaMA-3-7B, achieving a Modified BLEU score of 73.08 at 100\% data. However, \modelname{}, wo/g demonstrates a significant performance boost, surpassing all other methods with a score of 80.17 at 100\% data. For the \textsc{OpenExp} datasets, we notice that the performance of the LLaMA-3-7B and Mistral models is significantly lower than that of the T5-Base. We hypothesize that this phenomenon can be attributed to two aspects: 1) from the data aspect, the diversity of actions in \textsc{OpenExp} is greater compared to the \textsc{ChemTrans} dataset. When sampling 30\% of OPENEXP for training, there is a risk that certain actions may not be adequately sampled. This sampling issue poses a challenge for LLaMA-3 and Mistral, as these general-purpose LLMs struggle to perform optimally on skewed data distributions. Therefore, we may need more comprehensive data for supervised fine-tuning to eliminate the domain gap between general language tasks and domain-specific chemical knowledge. 2) From the metric aspect, we conduct a human evaluation experiment to evaluate the prediction results of all methods. The results demonstrate that Mistral achieves a score of 0.61, while T5-Base scored 0.50. These scores underscore Mistral's superior ability to align with human evaluators' expectations compared to T5-Base. Therefore, to some extent, Mistral doesn't exactly perform less well than T5-base.

\begin{table*}[ht]
\centering
\small
\renewcommand\arraystretch{1.08}
\resizebox{\linewidth}{!}{

\setlength{\tabcolsep}{0.7mm}{\begin{tabular}{l|cccc|cccc|cccc|cccc}
\toprule
\multicolumn{1}{l|}{\textbf{Metrics}}  & \multicolumn{4}{c|}{\textbf{90\% LEV}} & \multicolumn{4}{c|}{\textbf{75\% LEV}} & \multicolumn{4}{c|}{\textbf{Modified BLEU}} & \multicolumn{4}{c}{\textbf{Levenshtein similarity}}                                                              
\\ \midrule
\multicolumn{1}{l|}{\textbf{Training proportions}}   & \multicolumn{1}{c}{30\%} & \multicolumn{1}{c}{50\%} & \multicolumn{1}{c}{70\%} & \multicolumn{1}{c|}{100\%}  & \multicolumn{1}{c}{30\%} & \multicolumn{1}{c}{50\%} & \multicolumn{1}{c}{70\%} & \multicolumn{1}{c|}{100\%}  & \multicolumn{1}{c}{30\%} & \multicolumn{1}{c}{50\%} & \multicolumn{1}{c}{70\%} & \multicolumn{1}{c|}{100\%}  & \multicolumn{1}{c}{30\%} & \multicolumn{1}{c}{50\%} & \multicolumn{1}{c}{70\%} & \multicolumn{1}{c}{100\%} \\ \midrule
T5-Base        &0.00  & 0.13 & 0.38  & 1.01  & 0.89   & 3.42 & 6.58  & 13.16 & 52.37 & 59.25 & 64.38 & 68.94 & 46.19 & 51.02  & 54.33 & 58.75  \\ 
LLaMA-3-7B       &0.38  &0.51 & 0.51  & 0.38  & 4.94   & 6.71 & 5.32  & 7.97 & 47.89 & 51.12 & 49.22 & 52.92 & 46.65 & 48.51 & 46.86 & 49.45   \\
Mistral-7B       &\textbf{3.8} & \underline{4.56} & \underline{4.56}&\underline{5.44}   &\textbf{23.67} & \underline{26.33} &\underline{28.61} &\underline{30.51}  &\underline{70.05} &\underline{70.13} & \underline{71.51}&\underline{73.08}   &\underline{62.44} & \underline{63.83} &\underline{64.37}&\underline{65.52} \\
\textbf{ChemActor, wo/g}   &\underline{2.78}& \textbf{5.7} &\textbf{7.72} &\textbf{10.0} &\underline{22.53}& \textbf{29.62}&\textbf{33.8} &\textbf{39.75} &\textbf{72.83} &\textbf{77.03} &\textbf{79.32} &\textbf{80.17} &\textbf{63.75}  &\textbf{67.76} &\textbf{69.5} &\textbf{70.87}          
\\

\bottomrule
\end{tabular}}
}
\caption{Performance evaluation with varying different training proportions on the \textsc{ChemTrans} dataset.}
\label{tab:ChemTrans}
\end{table*}

\begin{table*}[ht]
\centering
\small
\resizebox{\linewidth}{!}{

\renewcommand\arraystretch{1.08}

\setlength{\tabcolsep}{0.6mm}{\begin{tabular}{l|cccc|cccc|cccc|cccc}
\toprule
\multicolumn{1}{l|}{\textbf{Metrics}}  & \multicolumn{4}{c|}{\textbf{90\% LEV}} & \multicolumn{4}{c|}{\textbf{75\% LEV}} & \multicolumn{4}{c|}{\textbf{Modified BLEU}} & \multicolumn{4}{c}{\textbf{Levenshtein similarity}}                                                             
\\ \midrule
\multicolumn{1}{l|}{\textbf{Training proportions}}   & \multicolumn{1}{c}{30\%} & \multicolumn{1}{c}{50\%} & \multicolumn{1}{c}{70\%} & \multicolumn{1}{c|}{100\%}  & \multicolumn{1}{c}{30\%} & \multicolumn{1}{c}{50\%} & \multicolumn{1}{c}{70\%} & \multicolumn{1}{c|}{100\%}  & \multicolumn{1}{c}{30\%} & \multicolumn{1}{c}{50\%} & \multicolumn{1}{c}{70\%} & \multicolumn{1}{c|}{100\%}  & \multicolumn{1}{c}{30\%} & \multicolumn{1}{c}{50\%} & \multicolumn{1}{c}{70\%} & \multicolumn{1}{c}{100\%}
\\ \midrule
T5-Base      &\underline{43.22}  & \underline{52.14} &\underline{56.29}  &\underline{63.02}  & \underline{79.95}   &  \underline{84.85} & \underline{86.77}  & \underline{89.85} & \underline{82.94} & \underline{85.39} & \underline{86.29} & \underline{88.37} & \underline{85.13} &  \underline{87.51} & \underline{88.58} & \underline{90.41}  \\ 
LLaMA-3-7B      & 13.83 & 13.25  & 13.95  & 16.08  & 51.66   & 50.72  & 51.95  & 55.01 &57.37 &  55.87 & 56.92 & 60.64 & 72.53 & 71.97  & 72.54 & 73.98  \\ 

Mistral-7B     & 38.56 & 41.7   & 43.42 & 43.98 & 75.81 & 77.62  & 78.7 & 78.88 & 78.67 & 79.97   & 80.47 & 80.31 & 83.16 & 84.05  & 84.58 & 84.63  \\
\textbf{ChemActor, wo/g}      &\textbf{64.2} &\textbf{65.59} &\textbf{68.31} &\textbf{69.4} &\textbf{89.72} &\textbf{89.98}   &\textbf{91.13} &\textbf{91.4} 
&\textbf{89.57} &\textbf{89.87}   &\textbf{91.05} &\textbf{92.9} 
&\textbf{90.47}  &\textbf{90.32}  &\textbf{91.55} &\textbf{91.63}  \\
\bottomrule
\end{tabular}}
}
\caption{Performance evaluation of ChemActor with varying different training proportions on the \textsc{OpenExp} dataset.}

\label{tab:openexp}
\end{table*}
\begin{table*}[ht!]
\centering
\small
\renewcommand\arraystretch{1.08}
\resizebox{\linewidth}{!}{

\setlength{\tabcolsep}{1mm}{\begin{tabular}{c|c|ccccc}
\toprule
\textbf{Model}    & \textbf{Method criterion}              & \textbf{SM-A} & \textbf{SM-O} & \textbf{BLEU-2} & \textbf{BLEU-4} & \textbf{EM}  \\ 
\midrule
\multirow{4}{*}{T5-Base} & 3k wo/generated                    & 41.55        & 66.26        & 45.19           & 30.07           & 2.15         \\ 
&  add 300 LLM-generated samples                & 44.38        & 67.87        & 47.34           & 31.91           & 3.29         \\ 
& add 600 LLM-generated samples                   & 54.36        & 75.88        & 52.78           & 37.17           &  8.99            \\ 
& add 2,100 LLM-generated samples                 &   57.71      &     77.9    &     55.08       &     39.24        &      10.76        \\ 
\midrule
\multirow{4}{*}{ChemActor} & randomly add 300 samples from \textsc{OpenExp}          & 67.10$\pm$0.03       & 83.29$\pm$0.05       & 74.17$\pm$0.05           & 66.32$\pm$0.06           & 30.19$\pm$0.08        \\ 
 & randomly add 600 samples from \textsc{OpenExp}          & 67.11$\pm$0.08        & 83.31$\pm$0.05        & 74.03$\pm$0.08           & 66.31$\pm$0.05           & 30.28$\pm$0.05        \\ 
& randomly add 2,100 samples from \textsc{OpenExp}         & 69.81$\pm$0.41        & 84.97$\pm$0.5        & 77.36$\pm$0.5           & 68.89$\pm$0.68           & 31.04$\pm$0.32        \\ 
& randomly add 50,000 samples from \textsc{OpenExp}   & \underline{71.08$\pm$0.8 }       & 85.31$\pm$0.8     & 78.48$\pm$0.5           & \underline{71.93$\pm$0.8 }          & 32.77$\pm$0.8\\
\midrule
\multirow{5}{*}{ChemActor}& 3k wo/generated  &68.29  & 85.07 & 75.63  &68.37 &31.9 \\
 & add 300 LLM-generated samples           & 68.29$\pm$0.07        & 84.97$\pm$0.05       & 76.17$\pm$0.03           & 68.74$\pm$0.05           & 32.41$\pm$0.1        \\ 
& add 600 LLM-generated samples           & 68.93$\pm$0.03        & 85.53$\pm$0.05       & 76.07$\pm$0.03           & 68.72$\pm$0.03           & 34.33$\pm$0.1        \\ 
& add 2,100 LLM-generated samples          & 70.88$\pm$0.04        & \underline{86.52$\pm$0.06}        & \underline{79.66$\pm$0.05}           & 69.67$\pm$0.08          & \underline{35.90$\pm$0.2}        \\ 
& add 50,000 LLM-generated samples  & \textbf{76.88$\pm$0.1}        & \textbf{89.01$\pm$0.06}       & \textbf{84.74$\pm$0.1}           & \textbf{76.93 $\pm$0.2}          & \textbf{36.4$\pm$0.4}        \\ 

\bottomrule
\end{tabular}}
}
\caption{Performance evaluation of ChemActor with varying amounts of LLM-generated data on the \textsc{ChemTrans} dataset. The best performance is in \textbf{bold}.}
\label{tab: add generated data}
\end{table*}

\subsubsection{Performance of LLM-Generated Data}
With the ascent of LLMs, natural language processing has seen improvements, including LLM-based data augmentation. However, an oversight arises from the random generation of augmented data by LLMs, suggesting that not all data may have equal training value, potentially impeding generative performance. We examine that \textit{``\textbf{how much will LLM-generated data benefit performance?}''} To answer this, we incorporate varying proportions of LLM-generated data into the training process and evaluate the model's performance on action-generation tasks. 


Table~\ref{tab: add generated data} presents the performance of ChemActor with varying amounts of LLM-generated data on the \textsc{ChemTrans} dataset. Method criterion refers to strategies for data augmentation. To assess the impact of our proposed data selection module, as illustrated in Figure~\ref{fig:framework}B, we also randomly sample data from the \textsc{OpenExp} dataset to construct training sets for training \modelname{}.

The results reveal that \modelname{}, when trained with an additional 50,000 LLM-generated samples achieves strong performance, attaining a BLEU-4 score of 76.93 and an EM score of 36.4, reflecting a 14.1\% improvement in EM. In contrast, incorporating the same volume of randomly sampled data from \textsc{OpenExp} leads to only a 2.7\% EM improvement. Additionally, when 300 randomly selected samples are added to the training set, the EM score decreases from 31.9 to 30.19, suggesting that indiscriminate data augmentation may introduce noise rather than improve model performance. In summary, our proposed data selection module is designed to identify reactions that differ from those in the original dataset, thereby expanding the representation of chemical reactions. This broader representation enhances the model’s ability to perform the D2A task more effectively.



\section{Conclusion}
In this paper, we present a fine-tuned LLM, a.k.a. ChemActor for automated extraction of chemical synthesis actions.
Trained on Q\&A instruction datasets with both real and LLM-generated data, ChemActor efficiently converts unstructured reaction descriptions into experimental actions. We introduce a data selection module that generates valuable data to address the uneven distribution of chemical space. Experimental results show that ChemActor achieves competitive results with state-of-the-art models, and our LLM-generated data proves more effective than other augmentation methods.

\section{Limitations}
Although ChemActor has been trained on extensive chemical reaction data, and we introduce a data selection module to tackle data sparsity issues, predicting actions with very limited observations remains a challenging task. In such cases, ChemActor might generate suboptimal answers. Furthermore, when applying ChemActor to a chemical automated platform, it's crucial to address potential safety risks from executing LLM-powered recommendations. To cope with this, developers must design comprehensive prompt guidelines that prioritize safety. If deemed dangerous, execution should be halted immediately, ensuring ChemActor’s application in automated platforms is secure and effective.

\section{Acknowledgments}
We thank SJTU AI for Science platform for the computing support. This work was supported by the Shanghai Municipal Science and Technology Major Project (2021SHZDZX0102), the Shanghai Municipal Science and Technology Explorer Project (24TS1403300), the National Natural Science Foundation of China (62102258), and the Fundamental Research Funds for the Central Universities.

\bibliography{custom}

\appendix

\section{Dataset Details}
\subsection{Data Collection} \label{A: dataset}
We utilize two datasets to evaluate the effectiveness of our method across description-to-action (D2A) and action-to-description (A2D) tasks.
\begin{itemize}
    \item \textbf{\textsc{OpenExp}~\cite{liu2024reactxt}.}
\textsc{OpenExp} is an open-source dataset that is derived from the raw data from USPTO-Applications~\cite{lowe1976chemical} and ORD~\cite{kearnes2021open}. It includes chemical reactions and the corresponding unstructured descriptions of experimental procedures. To obtain structured action sequences from reaction descriptions, authors conduct data preprocessing to filter low-quality data, leverage the action space defined by~\cite{vaucher2020automated}, and run the D2A model released by~\cite{christofidellis2023unifying}. The processed database features 274,439 pairs of chemical reactions and the corresponding descriptions and step-by-step actions for experimental procedures. It is randomly split into train/valid/test sets by the 8:1:1 ratio. 

We explore the inner distribution characteristics of the \textsc{OpenExp} dataset, which provides valuable insights into the data quality. Appendix Figure~\ref{fig:data-distribution}A-C displays the distribution of the number of actions, the number of description tokens, and the action taxonomy from the training and test set in the \textsc{OpenExp} dataset, respectively. It is indicated that the distribution of the number of actions is similar. Additionally, we can see that each data query comprises at least four actions and an average of ten actions (Appendix Figure~\ref{fig:data-distribution}A). Further, from Appendix Figure~\ref{fig:data-distribution}B, we can see that both of the distribution of the number of description tokens keeps a long-tail feature, and is characterized by an average token number of around 300. Appendix Figure~\ref{fig:data-distribution}C gives an idea of the frequency of action types in the dataset, which indicates that `ADD' and `STIR' account for the majority of the action space, while very few reactions involve `MICROWAVE' and `SONICATE'.

Further, we format the raw data from the \textsc{OpenExp} dataset to make it suitable to the D2A task. As the dataset is initially designed for the reaction-to-action (R2A) task, the molecules and the duration involved in the action are translated into tokens, such as `\$1\$' and `@5@'. To ensure that the action sentence is demonstrated in natural language, we revert the tokens to their original forms. However, most of the chemicals related to the `YIELD' action can not be mapped to their corresponding IUPAC names, which may cause confusion during the training period. As a result, we remove the `YIELD' action and the corresponding components for all experiments on the \textsc{OpenExp} dataset.

\item \textbf{\textsc{ChemTrans}~\cite{zeng2023transcription}.}
\textsc{ChemTrans} is an open-source human-annotated dataset sourced from the raw data from the supplementary information of Organic Syntheses. The database defines a concise and complete instruction schema for chemical synthetic actions, which contains 16 types of actions. It consists of 3,950 description-action pairs, on average with 154.6 tokens per input description and 9.2 actions per output action sequence. It is randomly split into training (2,765), validation (395), and test (790) sets. Additionally, by randomly sampling and substituting the operation sequences and corresponding arguments, T5-ChemTrans~\cite{zeng2023transcription} conducts data augmentation and expands the size of the training set to 29,326. To make a fair comparison to the T5-ChemTrans method, we employ the augmented dataset named as \textsc{ChemTrans}-30k for our experimentation.

To obtain more knowledge about the distribution characteristics of the training and test set from the \textsc{ChemTrans} dataset, we do an investigation on the distribution of the number of actions, the number of description tokens and the frequency of action types, as illustrated on Appendix Figure~\ref{fig:data-distribution}D-F. It is implied that most of the reactions contains nine actions (Appendix Figure~\ref{fig:data-distribution}D). Moreover, when the number of actions is fewer than five or more than fifteen, which is indicted in the two-side bars of the figure, the test set contains more instances with these action counts compared to the training set. As illustrated in Appendix Figure~\ref{fig:data-distribution}E, each reaction description consists of an average of 300 tokens, and very few descriptions contains more than 800 tokens. Thus, we set the maximum output text token length as 800 for the A2D task. Appendix Figure~\ref{fig:data-distribution}F shows that `ADD' and `SETTEMP' make up the majority of the action space. Besides, we can see that the distribution of the frequency of the action types from the training set is not very consistent to that from the test set. Actions such as `DISTILL' and `TRANSFER' occur much more frequently in the test set than in the training set, which poses a challenge on exactly deducing action types during the evaluation period.
\end{itemize}

\begin{figure*}[htp]
\centering 
\includegraphics[width=0.9\textwidth]{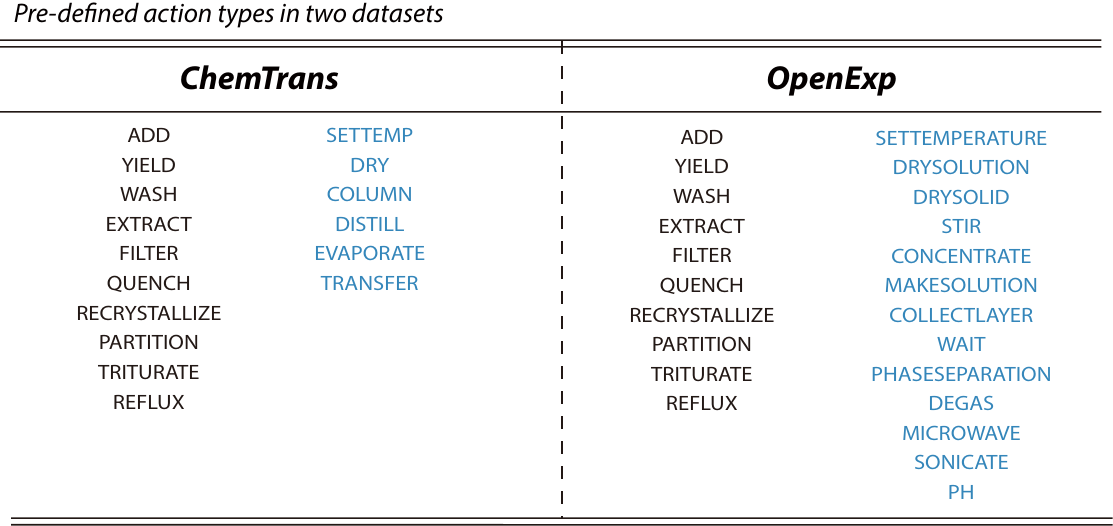}
\caption{Comparison of the action types on the \textsc{ChemTrans} and \textsc{OpenExp} datasets.}
\label{fig:action-type}
\end{figure*}

Apart from the two open-source description-action datasets mentioned above, there are some close-source action datasets. Vaucher et al.~\cite{vaucher2020automated} curate a database from the raw data from the commercial dataset Pistachio, and a number of studies employ it to perform the D2A task~\cite{vaucher2020automated, zhang2024fine, zhong2023reactie}. To make comparisons to the prior works, we do dataset statistics which is presented in Appendix Table~\ref{tab:case1}. Moreover, to obtain more insights into the pre-defined action space, Appendix Figure~\ref{fig:action-type} lists the action types annotated in the \textsc{ChemTrans} and the \textsc{OpenExp} dataset. The action types written in black appear in both datasets, while the action types written in blue only appear in one dataset. It is worth mentioning that though some action types are expressed in different forms, but they are similar in meaning. For instance, the `SETTEMP' action annotated in the \textsc{ChemTrans} dataset and the `SETTEMPERATURE' action annotated in the \textsc{ChemTrans} dataset share the same meaning. Additionally, the `DRY' action from the \textsc{ChemTrans} dataset is separated into the `DRYSOLUTION' and `DRYSOLID' actions from the \textsc{OpenExp} dataset. 

\begin{table*}[!t]
\centering
\small
\resizebox{0.8\textwidth}{!}{\begin{tabular}{l|ccccc}
\toprule
\multicolumn{1}{l|}{\textbf{Dataset}}   & \multicolumn{1}{c}{\textbf{Total}}  & \multicolumn{1}{c}{\textbf{Train}} &  \multicolumn{1}{c}{\textbf{Valid}}  &  \multicolumn{1}{c}{\textbf{Test}} & \multicolumn{1}{c}{\textbf{Open source}} \\ 
\midrule
Pistachio~\cite{vaucher2020automated}           & 693k & 555k  & 69k   & 69k   & No              
\\ 
\textsc{OpenExp}~\cite{liu2024reactxt}                   & 274k & 220k  & 27k & 27k   & Yes            
\\ 
\textsc{ChemTrans}~\cite{zeng2023transcription}          & 3,950 & 2,765 & 395    & 790   & Yes              
\\ 
\textsc{ChemTrans}-30k~\cite{zeng2023transcription}    & 30,511 & 29,326 & 395    & 790   & Yes              
\\
\bottomrule
\end{tabular}}
\caption{The statistics and comparison of different datasets.}
\label{tab:case1}
\end{table*}

\subsection{Construction of Pair-wised Instruction Datasets} \label{sec: data illu}
\begin{figure*}[t!]
\centering 
\includegraphics[width=1\textwidth]{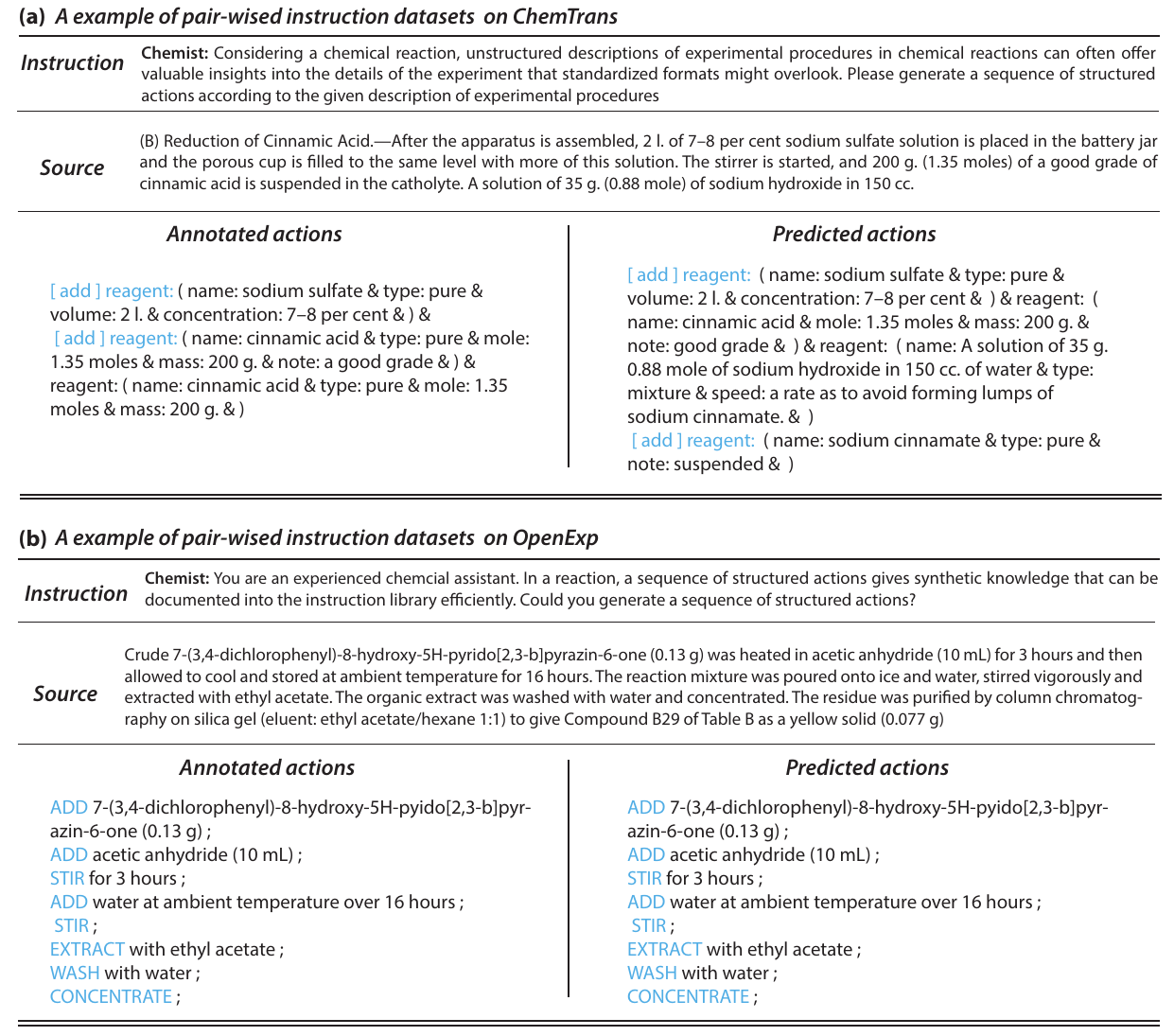}
\caption{Examples of constructed Q\&A datasets on \textsc{ChemTrans} and \textsc{OpenExp} datasets.}
\label{fig:data-example}
\end{figure*}

Instruction prompt datasets refer to formatting structured or unstructured data as natural language instructions, enabling LLMs to respond appropriately~\cite{reynolds2021prompt}. Recent advancements indicate that constructing high-quality prompt datasets facilitates effective reasoning of LLMs~\cite{wang-etal-2023-self-instruct}.
Towards the task of generating structural experimental actions from literature, we design a tailored instruction prompts system for better instruction tuning (Appendix Figure~\ref{fig:data-example}). Specifically, for both \textsc{ChemTrans} and \textsc{OpenExp} datasets, we first design instruction prompts for the definition of tasks, a.k.a. \textbf{\textit{Instruction}} in Appendix Figure~\ref{fig:data-example}. Next, we collect pair-wised reaction description-action sequences to construct high-quality Q\&A datasets. Question templates such as \textit{ `Please generate a sequence of structured actions according to the given description of experimental procedures'} are generated by GPT-4 autonomously using prompt engineering. `\textbf{\textit{Instruction}}' and `\textbf{\textit{Source}}' in Appendix Figure~\ref{fig:data-example} are integrated together as the input for instruction tuning. It is important to note that, we generate approximately 2,000 templates using GPT-4 to construct datasets, thereby ensuring the diversity and completeness of training sets. After constructing pair-wised Q\&A datasets, we further analyze the token distribution for two training sets, seen in Appendix Figure~\ref{fig:data-distribution}B. This indicates that the maximum text token length is 800, so we set it to 800 during training.


\begin{figure*}[htpb]
\centering 
\includegraphics[width=0.9\textwidth]{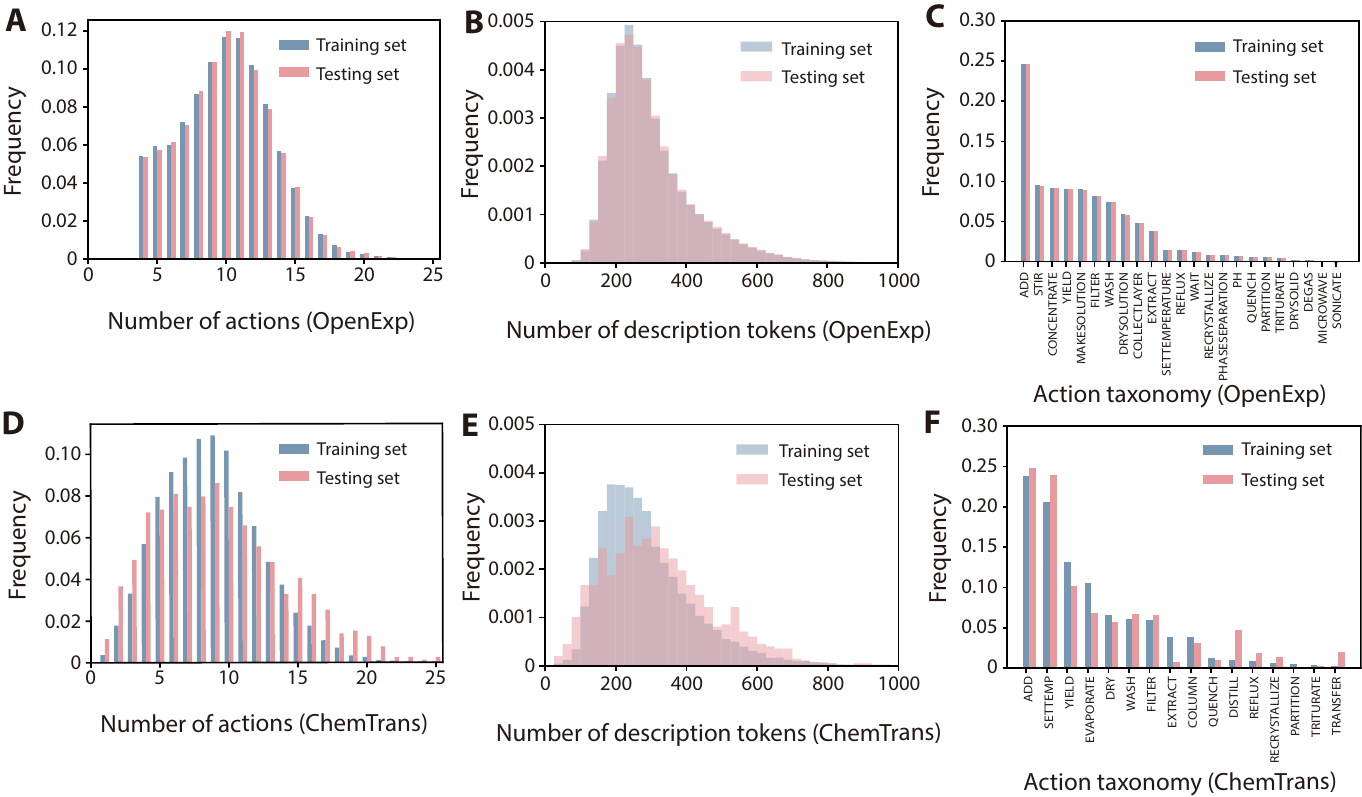}
\caption{Data distribution of the \textsc{OpenExp} and \textsc{ChemTrans} datasets.}
\label{fig:data-distribution}
\end{figure*}

\section{Experimental Details}\label{Experimental Details}
\subsection{Baseline methods}
We briefly introduce the baselines:
\begin{itemize}
    \item \textbf{Paragraph2Actions~\cite{vaucher2020automated}}. Paragraph2Actions is the first to propose the D2A task with pre-defined synthesis actions. Both Paragraph2Actions~\cite{vaucher2020automated} and Paragraph2Actions+~\cite{vaucher2021inferring} utilize a simple transformer-based model without any pre-training tasks to predict action sequences. Specifically, Paragraph2Actions+ incorporates additional knowledge-based enhancements to improve performance. Authors curate the data from the commercial Pistachio dataset, conduct data preprocessing, and collect description-action pairs. 
    
    \item \textbf{T5~\cite{raffel2020exploring}}. T5 is a method based on an encoder-decoder architecture, which has achieved competitive performance on a broad range of natural language processing tasks. For the T5-Base model, it consists of 12 encoder layers and 12 decoder layers, and is scaled up to 220M parameters.

    \item \textbf{T5-ChemTrans~\cite{zeng2023transcription}}. T5-ChemTrans utilizes a T5-based foundational model to facilitate transcription between human-readable reaction descriptions and machine-executable instructions. It is pre-trained on four knowledge enhancement tasks and fine-tuned on the augmented \textsc{ChemTrans} dataset.
    
    \item \textbf{GPT~\cite{achiam2023gpt}} GPT is a transformer-based large language model, which marks a massive leap in the process of language models. During its development, OpenAI enveils a series of GPT-based versions, including GPT-3.5-turbo and GPT-4o. In this work, we introduce GPT-3.5 to conduct D2A task. GPT-3.5 (text-davinci-003 completion mode) and GPT-3.5-chat (gpt-3.5-turbo chat completion mode) are employed for comparison. The original version displays the zero-shot performance. Further, GPT-3.5, 3-shot is given 3 randomly picked training instances, and GPT-3.5, 3-shot* is given 3 training instances with the highest similarity with the current testing instance. Additionally, we introduce GPT-4o for LLMs circle review. Compared to GPT-3.5-turbo, GPT-4o features a larger parameter scale and context window length, which improves its natural language processing capabilities.
    
    \item \textbf{ReactXT~\cite{liu2024reactxt}}. ReactXT is developed based on the MolCA model. It 
    incorporates three types of input contexts to incrementally pre-train LMs, which makes it suitable for experimental procedure prediction, retrosynthesis and molecule captioning tasks.
    
    \item \textbf{Mistral~\cite{jiang2023mistral}}. Mistral-7B-Instruct-v0.2 is a large language model with about 7.3B parameters. It trains transformers with grouped-query attention (GQA) and sliding window attention (SWA). 

    \item \textbf{LLaMA~\cite{llama-2}.} LLaMA is an open-source fine-tuned large language model, which has launched three versions so far. Compared to LLaMA-2, LLaMA-3 has made improvements on tasks such as chemistry-related questioning, code generation, and mathematics.


\end{itemize}
\subsection{Hyperparameters} \label{app:hyper}
\noindent

We offer a comprehensive summary of the training settings and hyperparameter values among ChemActor and baseline methods.

\begin{itemize}
    \item \textbf{ChemActor.} During the training process, we utilize the single-step retrosynthesis model proposed by~\cite{zeng2024ualign} to predict reactants given a product molecule. LLaMA-2~\cite{llama-2} is employed as our scientific foundational model. Fully fine-tuning LLaMA-2-7B for 8 epochs is completed in approximately 12 hours using 8 $\times$ NVIDIA A800 GPUs for \textsc{OpenExp} datasets. We set the batch size as 12, the default learning rate as 9.65e-6. 
    \item \textbf{T5-ChemTrans.} To perform the D2A and A2D tasks on the \textsc{ChemTrans} dataset, we utilize the T5-ChemTrans (Base) model checkpoints provided by~\cite{zeng2023transcription} and conduct generative inferences on the test set. Besides, for D2A tasks evaluated on the \textsc{OpenExp} dataset, we employ the source codes released by~\cite{zeng2023transcription} and fine-tune the T5-ChemTrans (Base) model on 220k description-action pairs from the \textsc{OpenExp} training set. During the fine-tuning period, we set the batch size to be 16, the learning rate to be 1e-4.
    \item \textbf{T5-Base.} We fine-tune the T5-Base model for performance improvement, which leverages the codes released by~\cite{zhang2024fine}. During the training period, we set the batch size as eight, the maximum number of epochs as five and the learning rate as 1e-5.
    \item \textbf{Mistral-7B-Instruct-v0.2.} We follow~\cite{zhang2024fine} to do full parameter fine-tuning for Mistral-7B-Instruct-v0.2. Our experimentation is capped for no more than three epochs, with the learning rate initialized as 1e-6 and the batch size as two.
    \item \textbf{LLaMA-3-7B-Instruct.} We also do full parameter fine-tuning for LLaMA-3-7B-Instruct. Our training settings are similar to those on the Mistral-7B-Instruct-v0.2 model.
    
\end{itemize}
\subsection{Metrics} \label{app:metric}

\textbf{Common Metric.} We assess performance using the Sequence Matching (SM) and ExactMatch (EM) metrics: SM-O evaluates the accuracy of operation prediction, while SM-A additionally considers the accuracy of argument prediction. For ExactMatch (EM), we report the proportion of items that are predicted perfectly. The detailed calculation of metrics is discussed in ~\cite{zeng2023transcription}.

Subsequently, following~\cite{vaucher2021inferring, liu2024reactxt}, we employ the additional metrics for performance evaluation in Table~\ref{tab:performance-openexp}, validity, which checks the syntactical correctness of the action sequence; the normalized Levenshtein similarity (LEV)~\cite{levenshtein1966binary}. Specifically, 90\%LEV refers to the proportion of predictions that achieve a normalized Levenshtein score greater than 0.9. The BLEU score~\cite{papineni2002bleu} and the ROUGE score~\cite{lin2004rouge}. BLEU focuses on precision, while ROUGE emphasizes recall in the generated text.  

\textbf{BERTScore.} Traditional natural language generation metrics assess the similarity between two strings by counting the minimum number of single-character edits (insertions, deletions, or substitutions) needed to transform one string into the other. These metrics cannot effectively capture the semantics of synthesis actions. To effectively evaluate \modelname{}, we introduce two additional metrics, including BERTScore~\cite{zhang2019bertscore} and multi-round LLMs circle review~\cite{chu2024pre}, shown in Figure~\ref{fig:framework}C-D. Considering BERTScore, it adopts encoder-only LLMs to evaluate the word embedding similarity of the prediction and the ground truth. BERTScore can adopt different LMs as the encoder, but the $deberta-xlarge-mnli$ model provides scores that are most closely aligned with human evaluations. Therefore, we select $deberta-xlarge-mnli$ to evaluate BERTScore, which can be computed by Appendix Equation~\ref{eq3}, where $\mathbf{E}_p = \{\mathbf{e}_p^1,\cdots, \mathbf{e}_p^k\}$ and $\mathbf{E}_t = \{\mathbf{e}_t^1,\cdots, \mathbf{e}_t^k\}$ are embeddings of the predictions and the ground truths, derived from the $deberta-xlarge-mnli$ model.

\begin{equation}\label{eq3}
    \left\{
\begin{array}{lll}
    R = \frac{1}{|\mathbf{E}_t|}\sum_{i=1}^{k} \max_{\mathbf{e}_p^j \in \mathbf{E}_p}\mathbf{e}_t^{i\top}{e}_p^{j}\\
    \\
    P = \frac{1}{|\mathbf{E}_p|}\sum_{j=1}^{k} \max_{\mathbf{e}_t^i \in \mathbf{E}_t}\mathbf{e}_t^{i\top}{e}_p^{j}\\
    \\
    F1 = 2\times \frac{P\times R}{P+R}
\end{array}
\right.
\end{equation}

\textbf{Multi-Round LLMs Circle Review}. Besides BERTScore, we ultilize the multi-round LLMs circle review to measure the semantic rationality of generated actions. We follow the debating strategy proposed by~\cite{chern2024can}. To be detailed, at round 0, we introduce a number of LLMs to evaluate the semantic similarity between the predicted and annotated actions. To standardize the evaluation results, we set a restriction that the LLM responses should output a score ranging from 0 to 1 and a sentence of explanation about why the score is marked. In this way, the evaluations from multiple LLMs are obtained at the initial stage. After that, we carry out a debate between LLMs. Specifically, at round N, we make the output scores and explanations generated by LLMs at round N-1 transparent to each other. Then, we apply a prompt-based method to create a debate among LLMs, which is illustrated in Appendix Figure~\ref{fig:debate}. Given the peer responses, each LLM will reconsider its previous response to the annotated and predicted actions, and provide a revision of its evaluation score and rationale. From the figure, we can see the score by GPT-3.5 shifts from 1.0 to 0.8 after the debate. After revising the responses on the whole dataset, the debate at round N is completed and the revised results will be spread to LLMs at round N+1. The debate will last for a fixed number of rounds, and the final average score is taken as the LLM score.

\section{Performance Evaluation}


\begin{table*}[!htbp]
\centering

\setlength{\tabcolsep}{5mm}{\begin{tabular}{cc|cc}
\toprule
\textbf{Action type} & \textbf{Recall (\%)} & \textbf{Action type} & \textbf{Recall (\%)} \\ 
\midrule
add & 85.82 & extract & 80.89\\
settemp &  89.35 & quench & 61.63 \\
yield & 79.15 & distill & 75.00 \\
evaporate & 76.48 & reflux & 55.71 \\
dry & 74.44 & recrystallize & 64.75 \\
wash & 70.12 & triturate & 71.43 \\
filter & 77.32 & transfer & 77.25 \\
column & 65.45 & \\
\bottomrule
\end{tabular}}
\caption{Results for action type predictions on \textsc{ChemTrans} dataset. }
\label{tab:action-type-recall}
\end{table*}


\begin{figure*}[t!]
\centering 
\includegraphics[width=0.85\textwidth]{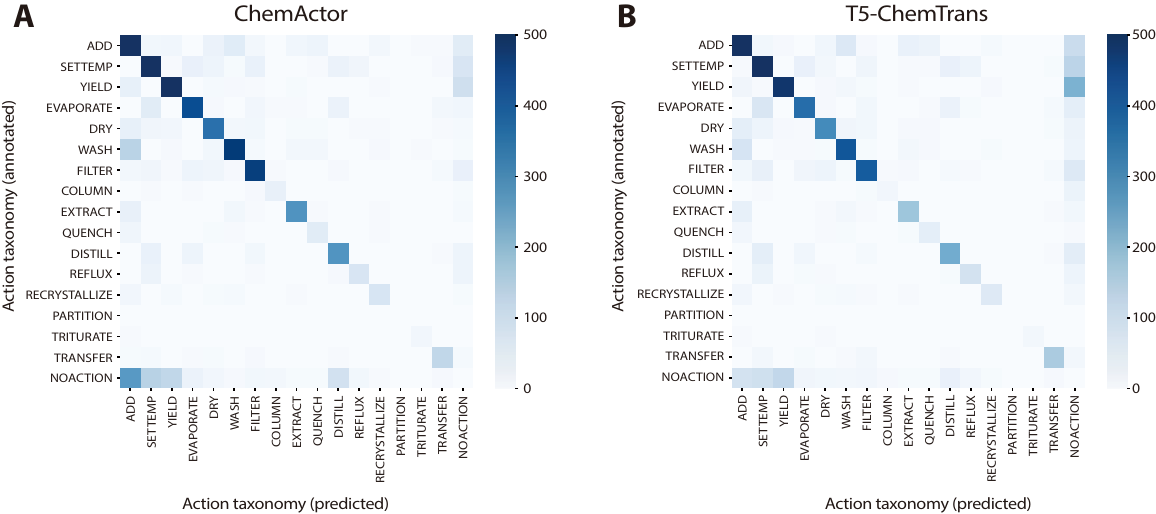}
\caption{Heatmap of type matching results of ChemActor and T5-ChemTrans.}
\label{fig:match-heat-map}
\end{figure*}

\subsection{Training Size Proportions} \label{app: training size}
To verify that the size of the training set can indeed improve the model performance, we adjust the proportions of training data on the \textsc{ChemTrans} and \textsc{OpenExp} datasets for the D2A task. Specifically, we randomly sort the training data and select the top data by a ratio of 0.1, 0.3, 0.5, 0.7, and 1.0, which forms a series of incremental data. In addition to our proposed ChemActor, we introduce T5-Base, Mistral-7B-Instruct-v0.2 and LLaMA-3, which are competitive D2A models and are discussed in the `Experimental Details' section. We adopt 95\% LEV, 75\% LEV, modified BLEU, and the average Levenshtein similarity to assess the model performance, which is introduced in the `Metrics' section. To be more intuitive, we visualize the performance of the Levenshtein similarity and the modified BLEU on the \textsc{ChemTrans} and \textsc{OpenExp} datasets, which are illustrated in Appendix Figure~\ref{fig:train-ratio}. From the figures, we can see that ChemActor performs consistently better than other methods on both evaluation metrics on the \textsc{ChemTrans} and \textsc{OpenExp} datasets, which highlights the effectiveness of our method.

\begin{figure*}[t!]
\centering 
\includegraphics[width=0.95\textwidth]{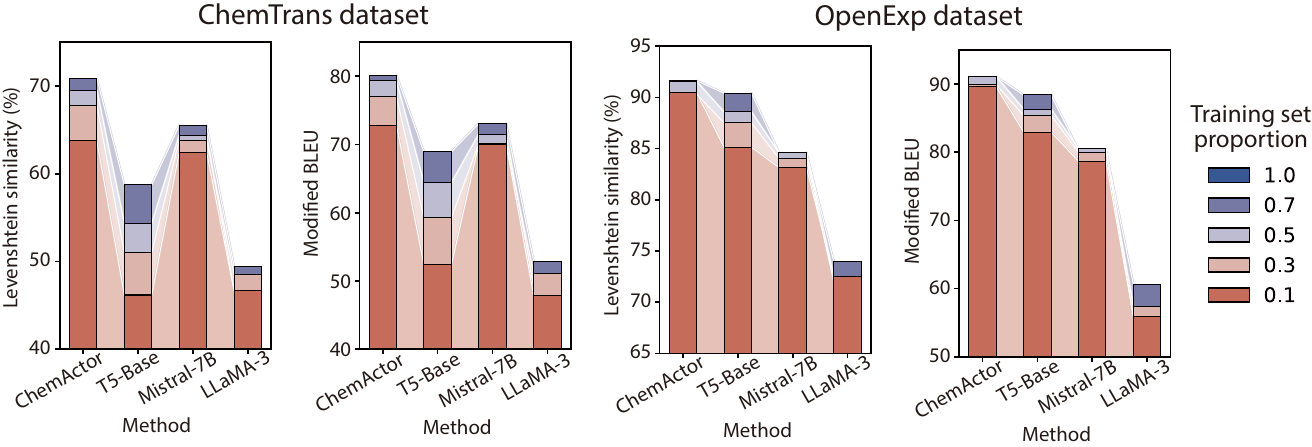}
\caption{The performance varying different training proportions on \textsc{ChemTrans} and \textsc{OpenExp} datasets.}
\label{fig:train-ratio}
\end{figure*}

The visualization reveals that the distribution of our proposed LLM-generated data has fewer overlaps with the distribution of real data. However, the distribution of augmented data shows more overlaps. By increasing the number of LLM-generated data, the distribution of our generated data gradually encompasses the entire chemical space. This indicates that the proposed data selection module is effective in generating diverse and useful data, which mitigates data sparsity issues during training and subsequently improves the performance of action sequence prediction.

\subsection{Type Matching} \label{app:type matching}


To evaluate the performance between our ChemActor and baseline methods, we employ the evaluation metrics mentioned in previous works, which have been introduced in the Section~\ref{app:metric}. However, among these metrics, the metrics derived from natural language tasks, such as BLEU and ROUGE, focus on the exact matching of sentences. Moreover, the metrics designed for the D2A task, such as SM-O, only care about the quality of action type classification and overlook the action components following the action type, which are indeed more important for chemical experimental procedures. Thus, in addition to these metrics, we further test type-matching experiments to evaluate the model's capability of perfectly predicting both the action types and the corresponding necessary components in the meantime.

To be specific, we first calculate the recall metrics of ChemActor in action type recognition, as presented in the Table~\ref{tab:action-type-recall}. From these results, we can observe that ChemActor exhibits varying recognition capabilities across different action types. For most action types, our model achieves a recall of over 75\%, demonstrating its excellent performance. However, for the action types ‘quench’ and ‘reflex’, the recall is only around 60\%. To investigate the reasons behind the model’s suboptimal performance on these two action types, we conduct an in-depth analysis of the action type distribution within the dataset. As shown in Appendix Figure~\ref{fig:data-distribution}, the action types ‘quench’ and ‘reflex’ appear with very low frequency in the training set, approximately 0.02. In that case, LLMs may struggle to learn effectively from extremely low-frequency samples, leading to the reduced recognition performance for these two action types. 

Further, we try to assess the model's ability to accurately predict both the types of actions and their corresponding essential components simultaneously. We separate the action sentence into a sequence of action phrases, and every action phrase contains one action type and a paragraph of action components. For example, the action phrase \textit{`[ quench ] reagent: ( name: ice water \& type: pure \& volume: 200 mL \& )'} comprises an action type \textit{`quench'} and the paired action components \textit{`reagent: ( name: ice water \& type: pure \& volume: 200 mL \& )'}. After extracting the action types and the action components from the predicted and annotated actions, we traverse the data to do action type matching. It is important to mention that instead of simply matching the predicted action types to the annotated action types, we compute the difflib similarity between every predicted-annotated action component pair. For each action phrase in the annotated action, we pick out the predicted action phrase containing the most similar action component to form a matching pair. To avoid mismatching, the matching is valid only when the similarity is higher than a certain threshold (we set it as 0.4), or its paired action phrase will be changed into `NOACTION'. For the action phrases in the predicted action, we follow similar steps. Finally, we count all the action type pairs and get the Type Matching results. In this way, the rule-based algorithm can do matches between the predicted and annotated action types that are most related to each other, thereby making the metric effectively evaluate the quality of generated actions.


We draw heatmaps of the type matching results, which are illustrated in Appendix Figure~\ref{fig:match-heat-map}. The evaluation results are generated by our ChemActor and a competitive method, T5-ChemTrans, which has been introduced in the `Experimental Details' section. In the figure, the labels of the predicted and annotated action types are on the x-axis and y-axis, correspondingly. The darker the color of the square is, the more pairs that satisfy the corresponding predicted and annotated action type. In this way, the darkness of the color of diagonal squares can effectively represent the model performance in making perfect action predictions. For the sake of clarity, the color scale is capped at 500, though many action-type pairs, particularly those on the diagonal, exceed this value. 

From Appendix Figure~\ref{fig:match-heat-map}, we can see that both ChemActor and T5-ChemTrans have a good performance on predicting action types, as the color of off-diagonal squares is much more shallow than that of diagonal squares. More importantly, the color of diagonal squares from ChemActor is greatly darker than that from T5-ChemTrans, which highlights the effectiveness of our method. Additionally, our proposed ChemActor presents an outstanding capability of learning from a few samples. For instance, the action type `DISTILL' occurs with a frequency of about 0.01 in the training set as depicted in Appendix Figure~\ref{fig:data-distribution}F, which highlights the challenge of representation learning. From Appendix Figure~\ref{fig:data-distribution}, we can see that the color of the `DISTILL-DISTILL' square in the left figure is much darker than that in the right figure, and the color of `DISTILL-X' and `X-DISTILL' square is much more shallow. It is indicated that ChemActor is less likely to make mistakes on the action `DISTILL' than T5-ChemTrans, which shows the great advantage of ChemActor on few-shot learning.

\subsection{Training stratiges}
\begin{figure}[ht]
\centering 
\includegraphics[width=0.465\textwidth]{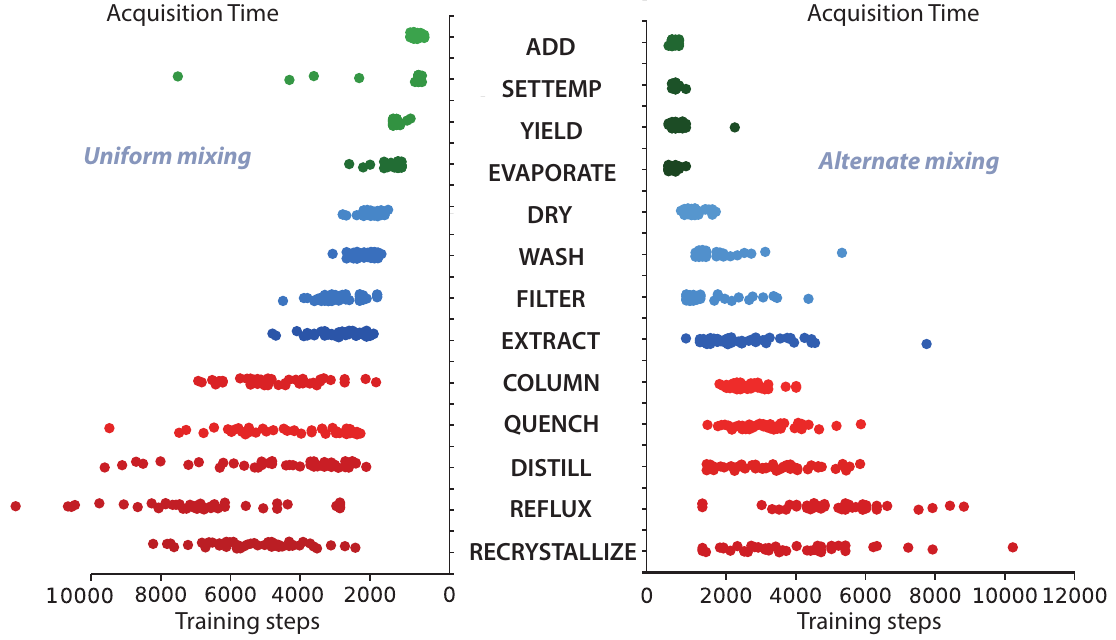}
\caption{The performance of different data curriculum methods. Acquisition time: The number of steps required to reach 90\% of final accuracy.}
\label{fig:skill speed}
\end{figure}

Various factors observed in the data may influence the speed of learning. We investigate the impact of LLM-generated data on model learning speed, focusing on data curriculum methods. Different skills require tailored training data, and the choice of data curriculum affects skill acquisition rates. Here, we define each action type in the \textsc{ChemTrans} datasets as a skill and measure the number of training steps needed to acquire these skills. We introduce two types of data curriculum methods, \textbf{uniform mixing}, where real and generated data are combined and fed to the model simultaneously, and \textbf{alternate mixing}, where real and generated data are fed in sequence, alternating during training. Instead of using next-word prediction loss, inspired by~\citep{evanson2023language}, we measure acquisition time, the number of steps needed for the model to reach 90\% of its final accuracy. Appendix 
Figure~\ref{fig:skill speed} discusses the relationship between data curriculum methods and model skills. All skills are sorted by the acquisition time.

From the results, we can see that different skills have different speeds of learning and emerge at different times. The smaller the percentage of actions in the training set, such as `DISTILL', and `REFLUX', the longer acquisition time it takes for the model to learn them. More importantly, the alternative mixing strategy enhances the speed of skill acquisition as demonstrated by the leftward movement of the data points in Appendix Figure~\ref{fig:skill speed}(right). This adjustment allows us to observe these points earlier in the training process.

\subsection{Multi-Round LLMs Circle Review} \label{app: circle review}

To evaluate our methods on the D2A task, we have applied a number of natural language metrics and proposed the Type Matching metric, which have been discussed in the previous sections. However, upon closely examining the predictions on the \textsc{ChemTrans} test set, we find that these metrics fail to comprehensively evaluate the accuracy and quality of the generated actions, as illustrated in Appendix Figure~\ref{fig:bad-case-1} and Appendix Figure~\ref{fig:bad-case-2}. To be specific, in Appendix Figure~\ref{fig:bad-case-1}, the action generated by our ChemActor paraphrases the action phrase \textit{`[ wash ] reagent: ( name: two 60-mL portions of a 1:1 mixture of saturated aqueous sodium bicarbonate and water \& type: mixture \& volume: 60-mL \& concentration: 1:1 \& batch:each: portions \& )'} as \textit{` [ wash ] reagent: ( name: sodium bicarbonate \& type: mixture \& volume: 60-mL \& concentration: saturated \& note: aqueous \& batch:each: portions \& ) \& reagent: ( name: water \& type: pure \& )'}. It can be inferred that the predicted actions are semantically similar to the annotated actions. However, the generated actions are not in the same data format or in the same order as the human annotations, which leads to a low difflib similarity of 13.18. In that point, the previous metrics which only focus on exact matching fail to handle the diversity of action outputs, and underestimate the effectiveness of our model. Thus, it is crucial to apply an evaluation metric capable of effectively measuring semantic rationality. 

We apply the multi-round LLMs circle review to evaluate the semantics of synthesis actions, and the details of the metric are introduced in the Appendix Section~\ref{app:metric}. During our experimentation, to evaluate the effectiveness of our method on the D2A task, we introduce Mistral-7B-Instruct-v0.2, T5-Base, and T5-ChemTrans as baseline methods, whose experimental details are discussed in the Appendix Section~\ref{Experimental Details}. We curate a bad case dataset and conduct the LLM-based multi-round circle review evaluation. Specifically, we traverse the actions predicted by our ChemActor and other methods and compute the difflib similarity between the ground truth and the prediction. Then, we analyze the data to filter cases with a difflib similarity greater than 0.4, remaining 279 bad cases. Additionally, we adopt GPT-3.5-turbo, GPT-4o, LLaMA-3, and Mistral-7B-Instruct-v0.2, which show great competitiveness in natural language processing tasks. Further, we set the number of debate round as two and conduct the multi-round LLMs circle review evaluation.


The evaluation results are illustrated in Figure ~\ref{fig:peer review}. Our method outperforms other baselines on the multi-round circle review score based on all LLMs. To be more detailed, we give two examples of multi-round LLMs circle review evaluation on bad cases, which are illustrated in Appendix Figure~\ref{fig:bad-case-1} and Appendix Figure~\ref{fig:bad-case-2}. We can see that the results generated by our method are semantically similar to the ground truth, and achieves a relatively low difflib similarity score but the highest LLM review score. It implies that the difflib similarity metric simply measures the sentence similarity token by token, which may overlook the diversity of action predictions and underestimate the predictions generated by our method. However, the multi-round LLMs circle review is able to evaluate the semantic rationality and provide a more credible evaluation result.


\begin{figure*}[htp]
\centering 
\includegraphics[width=0.9\textwidth]{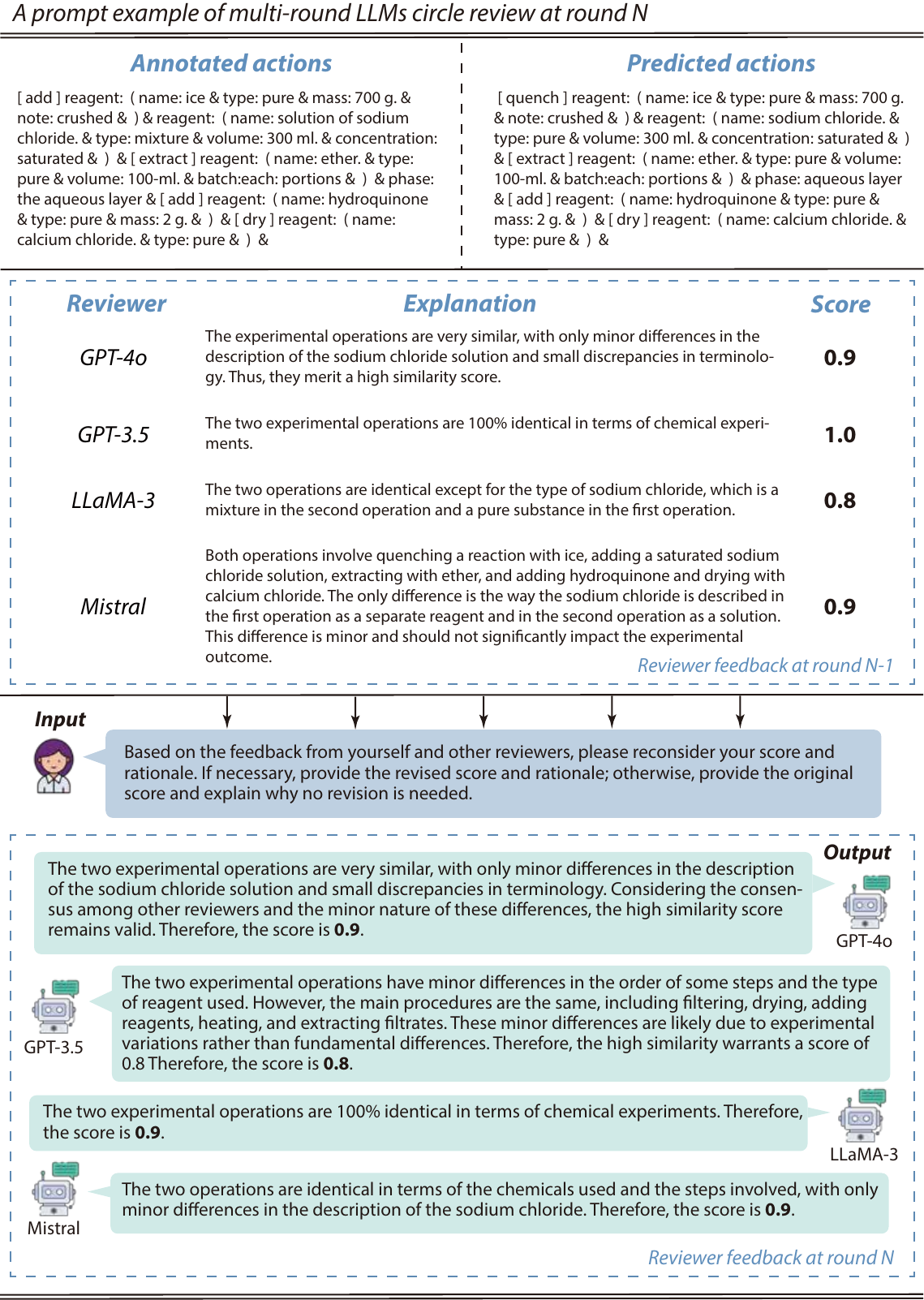}
\caption{The pipeline of the multi-round LLMs circle review.}
\label{fig:debate}
\end{figure*}

\begin{figure*}[htp]
\centering 
\includegraphics[width=0.95\textwidth]{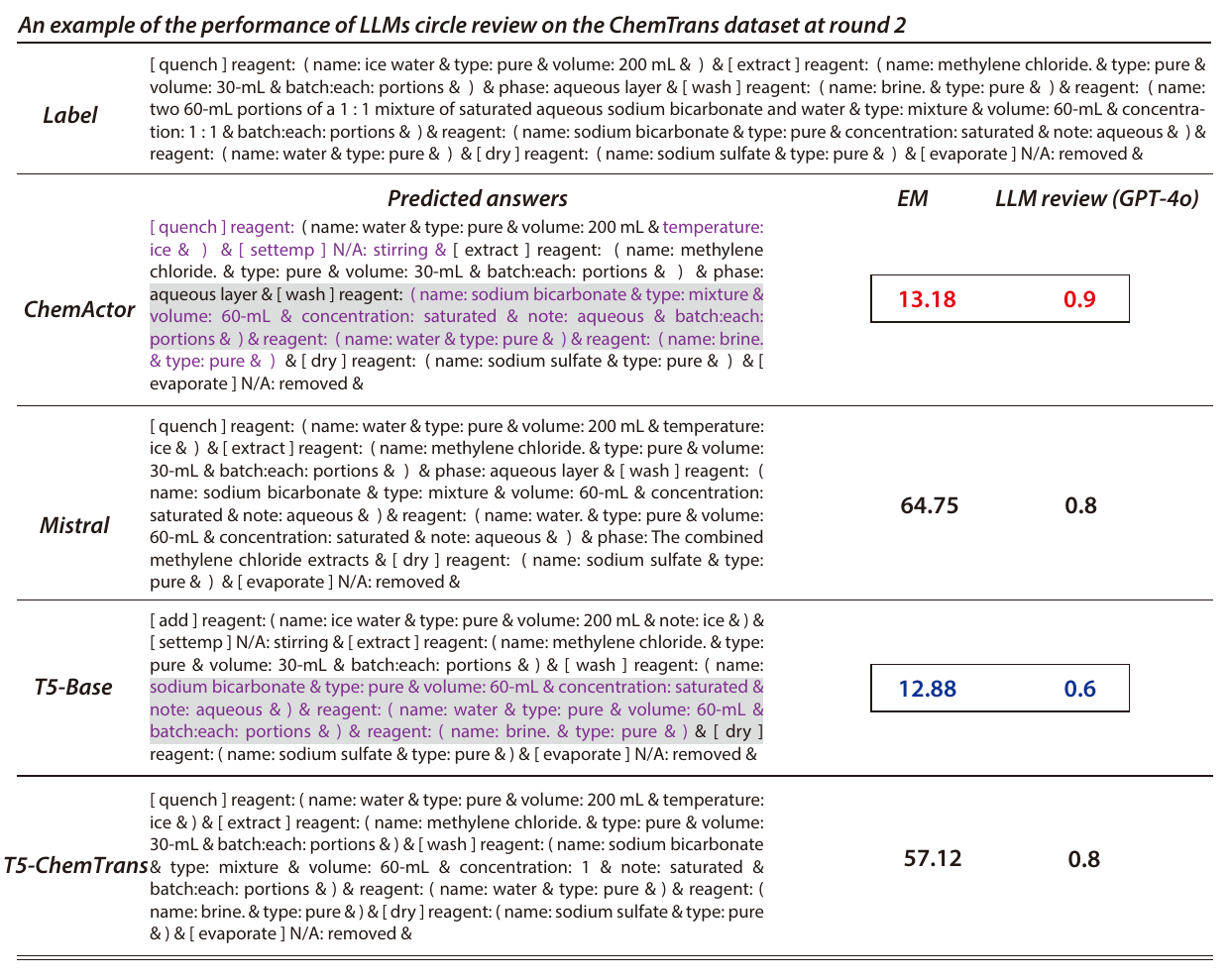}
\caption{An example of the performance of multi-round LLMs circle review (bad case 1).}
\label{fig:bad-case-1}
\end{figure*}

\begin{figure*}[htp]
\centering 
\includegraphics[width=0.95\textwidth]{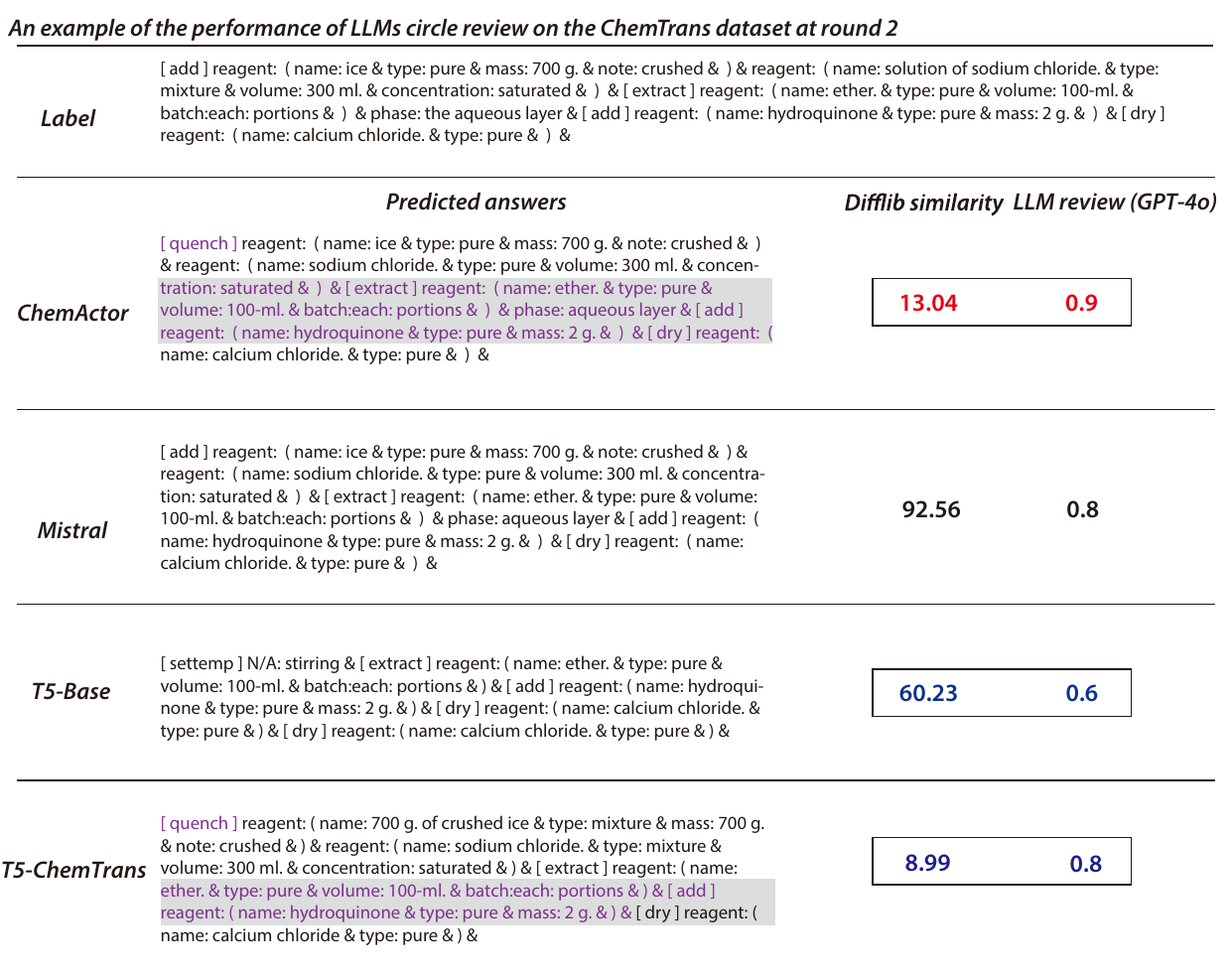}
\caption{An example of the performance of multi-round LLMs circle review (bad case 2).}
\label{fig:bad-case-2}
\end{figure*}

\begin{figure*}[htp]
\centering 
\includegraphics[width=0.95\textwidth]{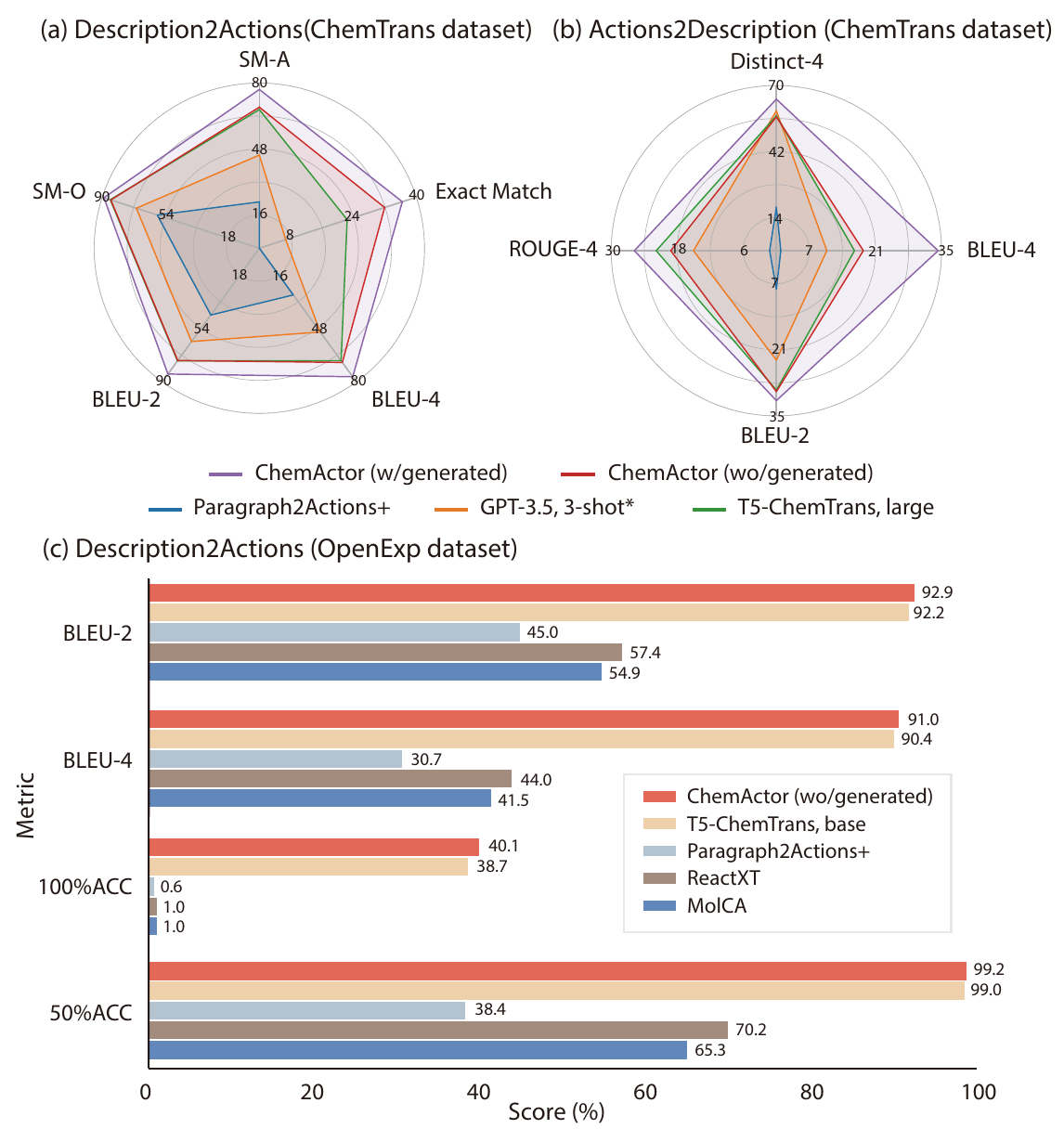}
\caption{Experimental results of ChemActor on \textsc{ChemTrans} and \textsc{OpenExp} datasets.}
\label{fig:baseline-result}
\end{figure*}

\end{document}